\documentclass[10pt,twocolumn,letterpaper]{article}

\usepackage[pagenumbers]{cvpr} % To force page numbers, e.g. for an arXiv version

\usepackage{amsmath}
\usepackage{amssymb}
\usepackage{booktabs}
\usepackage{multirow}
\usepackage{makecell}

\definecolor{cvprblue}{rgb}{0.21,0.49,0.74}
\usepackage[pagebackref,breaklinks,colorlinks,allcolors=cvprblue]{hyperref}

\def\paperID{*****} % *** Enter the Paper ID here
\def\confName{CVPR}
\def\confYear{2026}

\title{Multi-Modal Building Inspection via Perceiver IO Fusion\\of Satellite and Street-Level Imagery}

\author{Niels Sombekke$^{1}$\quad
Rob G.J. Wijnhoven$^{2}$\quad
Martin R. Oswald$^{1}$\\
$^{1}$University of Amsterdam (UvA), Amsterdam, The Netherlands\quad
$^{2}$Spotr, The Hague, The Netherlands\\
{\tt\small nielssombekke@gmail.com\quad rob@spotr.ai\quad m.r.oswald@uva.nl}
}

\begin{document}
\maketitle

\begin{abstract}
We present a multi-modal classification framework that fuses satellite and street-level imagery through a Perceiver IO architecture operating on spatial patch tokens from a shared DINOv2 backbone.
The design naturally handles a variable number of street-level views per building without padding or fixed-size pooling, and jointly predicts multi-label roof element and roof material classes.
We construct a large-scale dataset of 32,135 buildings (61,672 segments) spanning ten countries, pairing satellite images with up to eight street-level views per segment and evaluating four masking strategies for isolating the target building.
We propose an RGB-M masking strategy that appends the building footprint mask as a fourth input channel, providing a soft spatial prior that outperforms hard cropping across both modalities.
The Perceiver IO fusion model improves over all other fusion strategies and yields substantial per-class gains for attributes visible from street level (e.g., +11.3 AP for slate, +1.3 AP for dormers), though the satellite-only baseline retains a slight advantage in macro-averaged mAP for classes that are predominantly visible from above.
These results establish a scalable, flexible architecture for multi-modal building inspection that can accommodate heterogeneous inputs and multiple output tasks.
\end{abstract}

\section{Introduction}
\label{sec:intro}

\begin{figure}[t]
  \centering
  \includegraphics[width=\linewidth]{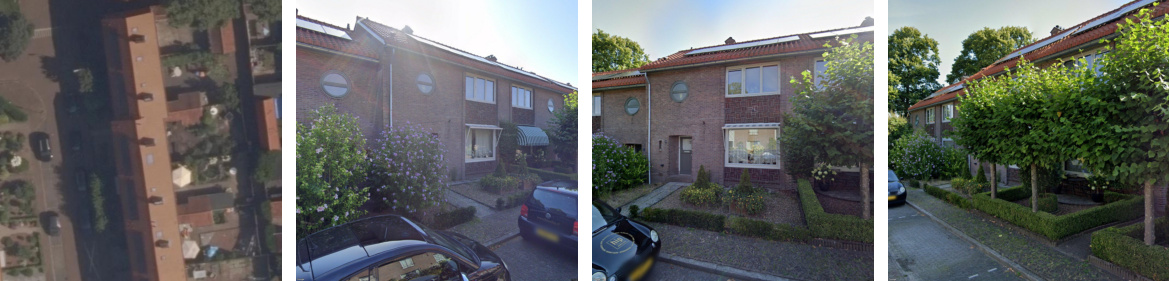}
  \caption{Example input data for a building: satellite image (left) and corresponding street-level images (right). Satellite imagery captures roof geometry and context, while street-level views reveal facade details, materials, and structural elements that are invisible from above.}
  \label{fig:teaser}
\end{figure}

Maintaining and insuring the built environment at national scale requires detailed knowledge of individual buildings: what materials cover the roof, which structural elements are present, and what condition they are in. Traditional inspection relies on labor-intensive site visits that are costly, slow, and prone to human error, while generating significant CO\textsubscript{2} emissions from repeated travel~\cite{rundle2011using}. With the European building stock alone comprising over 200 million structures, 40\% of the EU's energy consumption and 36\% of its greenhouse gas emissions~\cite{eu_buildings}, manual approaches simply cannot keep pace with the demand for up-to-date property information. A scalable, data-driven alternative is needed.

Remote sensing offers a scalable path. Satellite and aerial imagery can be collected over large areas, updated regularly, and processed algorithmically, enabling the analysis of thousands of properties without physical visits. Satellite data captures a building's footprint, roof geometry, and surrounding context from above, while street-level imagery---increasingly available through services such as Google Street View---provides close-range views of facades, wall conditions, and material textures that are invisible from a top-down perspective. Combining both modalities can produce richer, multi-perspective building profiles than either source alone~\cite{cao2018integrating, hoffmann2019model, law2019take}.

However, fusing satellite and street-level imagery poses several challenges. First, the data itself is heterogeneous: satellite images may have limited resolution and suffer from misaligned building footprints, while street-level views are frequently occluded by vegetation, vehicles, or neighboring structures~\cite{he2021urban}. Second, an effective method must isolate the target building from its surroundings---leveraging the building footprint to focus on the relevant structure---while retaining enough environmental context to support classification. Third, the number of available street-level images varies widely across buildings: some have dozens of views from different angles, others have none at all. A practical architecture must therefore handle a variable number of inputs without degrading performance on buildings with sparse or missing street-level coverage.

We address these challenges with a Perceiver IO-based fusion architecture~\cite{jaegle2021perceiver} that operates directly on spatial patch tokens extracted by a shared DINOv2~\cite{oquab2023dinov2} backbone. Unlike feature-vector fusion approaches that discard spatial structure through global pooling, our model preserves fine-grained patch-level information from both modalities and compresses them through cross-attention into a fixed-size latent representation. The architecture naturally accommodates varying numbers of street-level inputs through its set-based cross-attention mechanism, and jointly predicts multi-label roof element and roof material classes via task-specific output heads.

We make the following contributions:
\begin{enumerate}
    \item We develop a Perceiver IO fusion model that processes spatial tokens from satellite and street-level imagery, outperforming concatenation and transformer-based fusion baselines and delivering substantial gains for classes where street-level views are informative.
    \item We systematically evaluate four masking strategies for isolating the target building and show that an RGB-M approach, appending the binary footprint mask as a fourth channel, provides the best trade-off between spatial focus and contextual information.
    \item We construct a multi-modal building inspection dataset of 32,135 buildings (61,672 segments) across ten countries, pairing satellite imagery with variable numbers of street-level views and building footprint masks.
    \item We compare two backbones (ResNet-50, DINOv2-S) and multiple fine-tuning strategies, demonstrating that fully fine-tuned DINOv2-S consistently outperforms the convolutional baseline on both modalities.
\end{enumerate}

\section{Related Work}
\label{sec:related}

\paragraph{Satellite imagery for building analysis.}
CNN-based methods have been widely applied to satellite imagery for building footprint extraction~\cite{ji2018fully, zhao2018building, ji2019scale}, damage detection~\cite{xu2019building}, and roof type identification~\cite{kim2021cnn}. Vision Transformers have since improved large-scale building extraction~\cite{wang2022building} and damage assessment~\cite{kaur2023large} through their ability to capture global spatial context. More recently, self-supervised foundation models such as DINOv2~\cite{oquab2023dinov2} have demonstrated strong transfer to remote sensing tasks, particularly in low-label regimes~\cite{lu2025vision, scheibenreif2022self}. Despite these advances, relying solely on satellite imagery limits inspection to attributes visible from above, missing facade-level detail that is critical for comprehensive building assessment.

\paragraph{Street-level imagery for building analysis.}
Street-level images offer ground-level perspectives on facades, materials, and structural elements that are occluded in satellite views. Deep learning approaches have been applied to building typology classification~\cite{gonzalez2020automatic, taoufiq2020hierarchynet, laupheimer2018neural}, property value estimation~\cite{zhao2018deep}, facade parsing~\cite{kong2020enhanced, wang2024improving}, and floor count estimation~\cite{iannelli2017extensive}. Semi-supervised approaches combine street-level imagery with auxiliary data such as OpenStreetMap metadata for building height estimation~\cite{li2023semi}. Biljecki and Ito~\cite{biljecki2021street} provide a comprehensive survey of street-view imagery in urban analytics. Key challenges include variable lighting and weather conditions, occlusion by vegetation and vehicles, inconsistent camera calibration across regions, and privacy-related blurring~\cite{he2021urban}.

\paragraph{Fusion of satellite and street-level modalities.}
Combining satellite and street-level data leverages their complementary strengths: broad spatial coverage from above and detailed facade information from ground level. This multi-modal approach has improved performance in house price estimation~\cite{law2019take}, land use classification~\cite{cao2018integrating, srivastava2019understanding}, building type classification~\cite{hoffmann2019model}, socioeconomic prediction~\cite{suel2021multimodal}, and flood risk assessment~\cite{xing2023flood}. Early fusion methods concatenate or sum global feature vectors, treating modalities symmetrically without exploiting their complementary spatial structure~\cite{fan2022multilevel}. Recent attention-based approaches process multiple views with learned weighting, focusing on the most informative inputs~\cite{chen2022multi, huang2023comprehensive, guo2024fusion}. However, most existing methods operate on a fixed number of views per building, discarding spatial token-level detail through global pooling before fusion.

Our work differs in three respects. First, we fuse at the \emph{spatial token level} rather than the feature vector level, preserving fine-grained patch information from both modalities. Second, we employ Perceiver IO~\cite{jaegle2021perceiver} to compress a variable-length token sequence into a fixed-size latent representation through cross-attention, naturally handling buildings with any number of street-level views. Third, we build on a self-supervised DINOv2 backbone rather than task-specific supervised features, providing richer representations for both modalities.

\section{Method}
\label{sec:method}

We describe the dataset construction (\cref{sec:dataset}), masking strategies for isolating the target building (\cref{sec:masking}), uni-modal baselines (\cref{sec:unimodal}), and the Perceiver IO fusion architecture (\cref{sec:fusion}).

\subsection{Dataset}
\label{sec:dataset}

\begin{figure}[t]
  \centering
  \includegraphics[width=\linewidth]{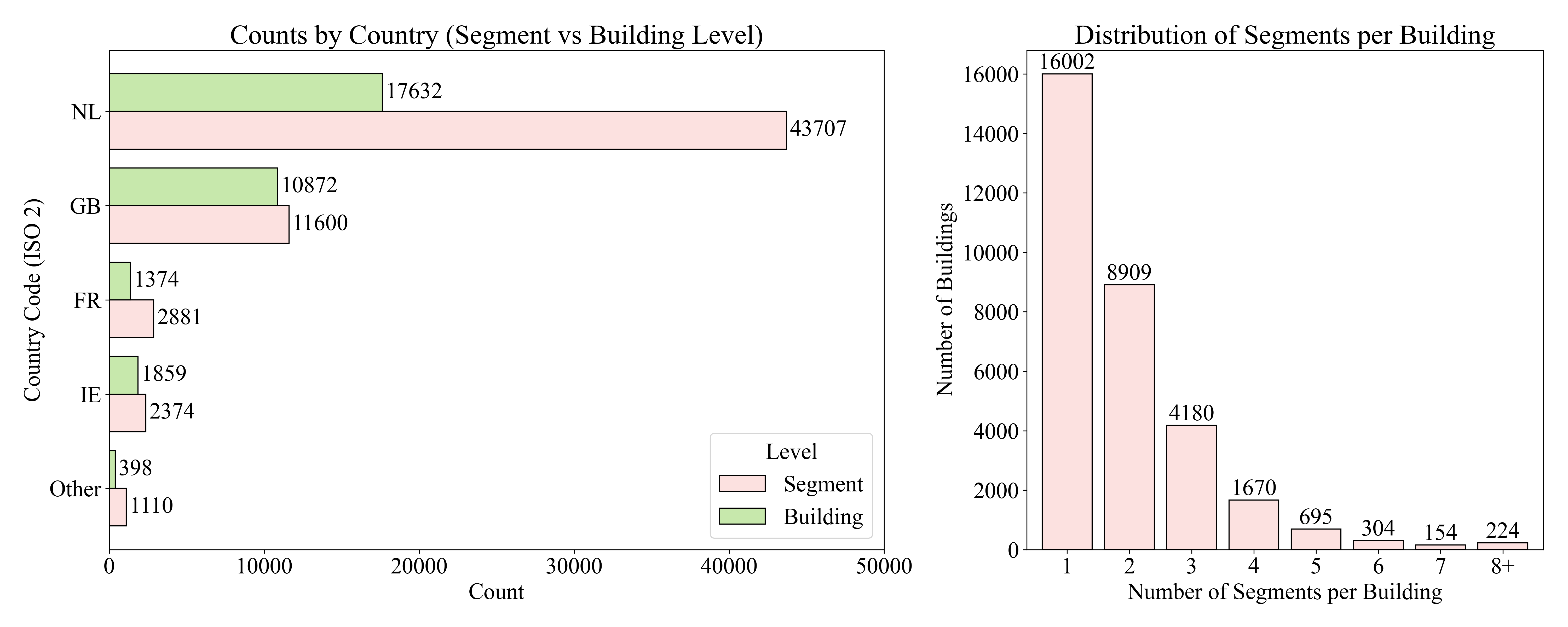}
  \caption{Dataset statistics: building and segment counts by country (left) and distribution of segments per building (right). The Netherlands contributes the largest share.}
  \label{fig:dataset-stats}
  \vspace{2mm}
\end{figure}

We construct a multi-modal dataset of 32,135 buildings (61,672 segments) across ten countries, with the Netherlands (17,632), UK (10,872), and France (1,374) contributing the largest shares. Labels are assigned at the roof segment level, where each building may contain multiple roof segments. The dataset supports two multi-label classification tasks: \textbf{roof elements} (6 classes: solar panels, dormer, skylight, roof window, chimney, external installations) and \textbf{roof materials} (7 classes: bitumen, slate, tiles, aluminium, thatch, corrugated sheets, glass). Class distributions are heavily imbalanced (\cref{tab:label-count}), with chimney (19,155 segments) and bitumen (27,293) being the most frequent, while thatch (184) and glass (280) are extremely rare. \Cref{fig:dataset-stats} shows the geographic distribution of the dataset.

\begin{table}[t]
\centering
\caption{Label distribution at segment level. Both tasks exhibit strong class imbalance.}
\label{tab:label-count}
\small
\begin{tabular}{llr}
\toprule
Task & Label & Count \\
\midrule
\multirow{6}{*}{Roof Elements}
& Solar Panels           & 6,905 \\
& Dormer                 & 4,210 \\
& Skylight               & 999 \\
& Roof Window            & 10,726 \\
& Chimney                & 19,155 \\
& External Installations & 2,064 \\
\midrule
\multirow{7}{*}{Roof Materials}
& Bitumen                & 27,293 \\
& Slate                  & 6,932 \\
& Tiles                  & 23,874 \\
& Aluminium              & 2,693 \\
& Thatch                 & 184 \\
& Corrugated Sheets      & 1,811 \\
& Glass                  & 280 \\
\bottomrule
\end{tabular}
\end{table}

\begin{figure}[t]
  \centering
  \includegraphics[width=\linewidth]{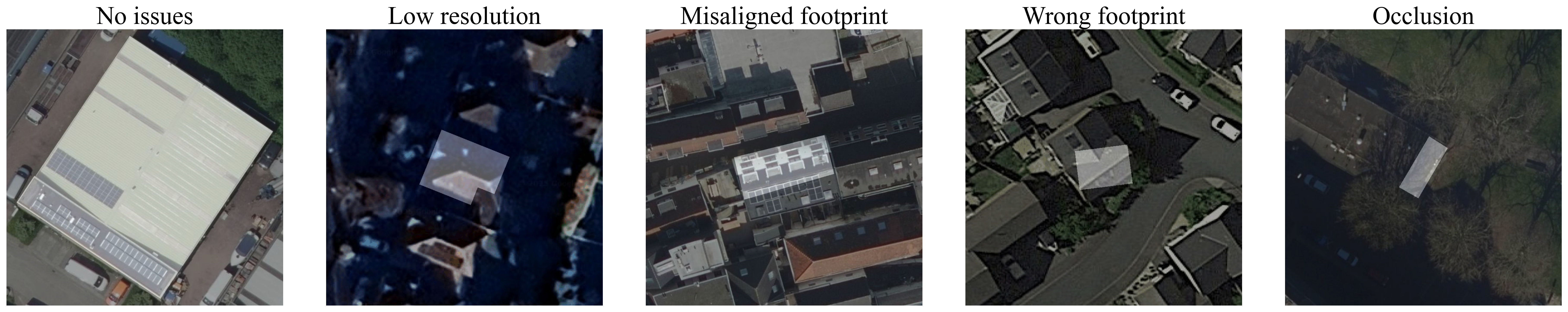}
  \caption{Satellite imagery challenges: no issues, low resolution, misaligned footprint, wrong footprint, and occlusion by trees/shadows. Showing image and building mask.}
  \label{fig:sat-challenges}
  \vspace{2mm}
\end{figure}

\paragraph{Satellite imagery.} For each segment, a single satellite image is cropped to a fixed 42$\times$42\,m physical area centered on the building. Sources include Google Maps and higher-resolution national aerial imagery (e.g., PDOK in the Netherlands). Building footprints are obtained from OpenStreetMap~\cite{OpenStreetMap} or the Dutch BAG cadastral registry (\cref{fig:sat-challenges}).

\begin{figure}[t]
  \centering
  \includegraphics[width=\linewidth]{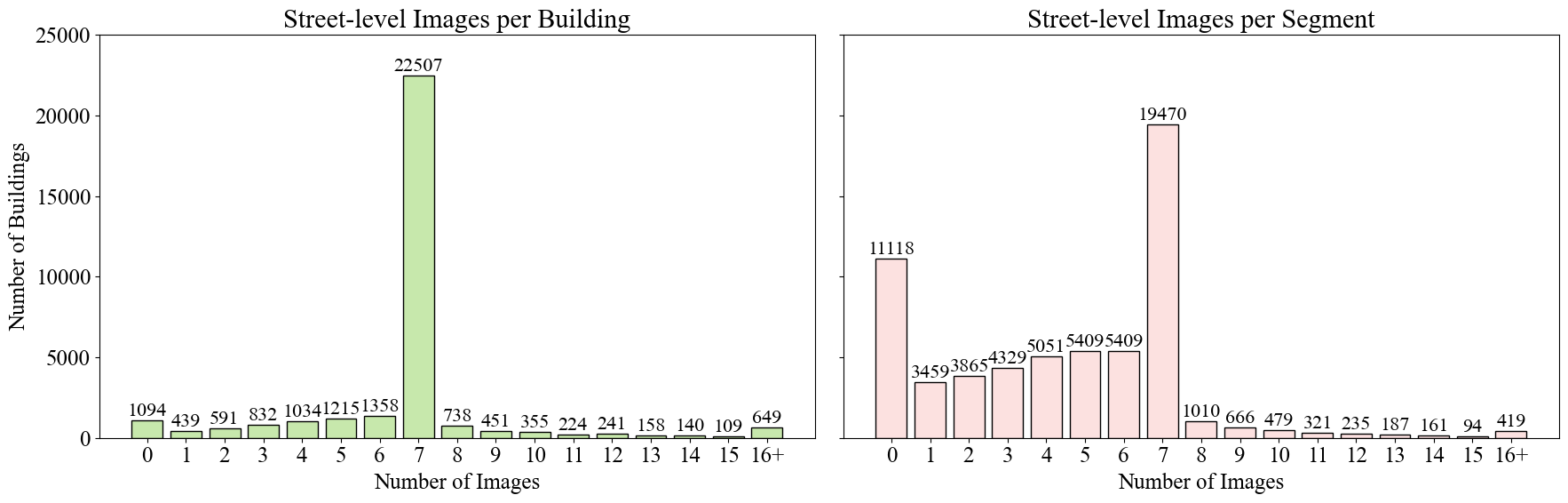}
  \caption{Distribution of street-level images per building (left) and per segment (right). Most buildings have 7 images; at the segment level, many segments have 0 or 7 images.}
  \label{fig:sl-stats}
  \vspace{2mm}
\end{figure}

\begin{figure}[t]
  \centering
  \includegraphics[width=\linewidth]{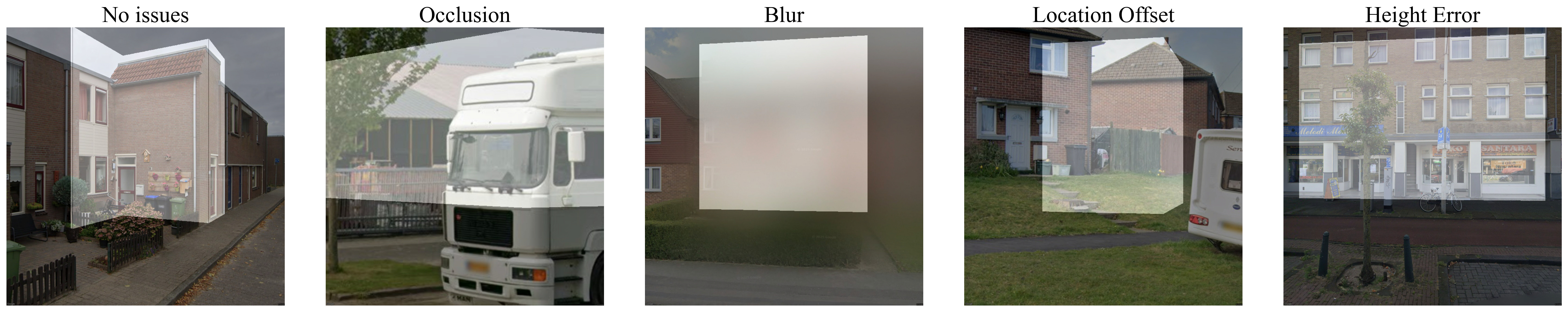}
  \caption{Street-level imagery challenges: no issues, occlusion by vehicles/vegetation, privacy blurring, camera location offset, and building height error affecting mask projection.}
  \label{fig:sl-challenges}
  \vspace{2mm}
\end{figure}

\begin{figure}[t]
  \centering
  \includegraphics[width=\linewidth]{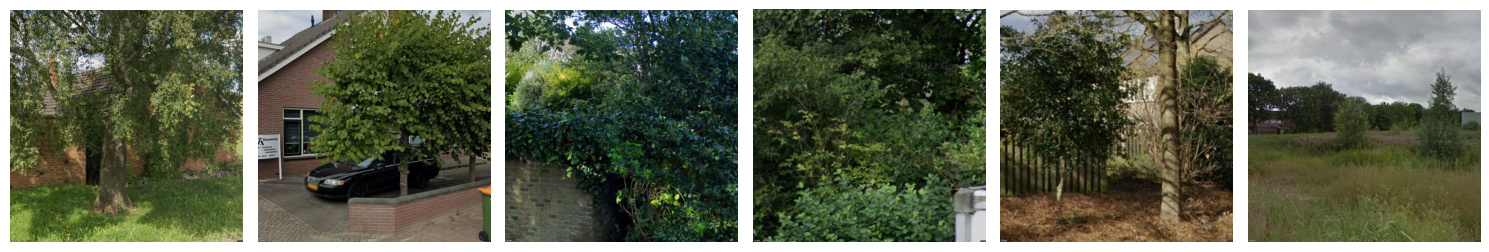}
  \caption{Examples of street-level occlusion patterns. Vegetation, vehicles, and neighboring structures partially obscure the target building.}
  \label{fig:sl-occlusion}
  \vspace{2mm}
\end{figure}

\paragraph{Street-level imagery.} Street-level views are sourced from Google Street View and, in the Netherlands, from Cyclomedia panoramas. A custom selection algorithm maximizes facade visibility while minimizing redundancy, yielding an average of 4.54 images per segment (280,150 image-mask pairs total). For each image, a pixel-level mask is generated by projecting the building footprint into the camera view using the footprint polygon, camera position, and an estimated building height. Images with less than 20\% building visibility (as assessed by a SegFormer-B0~\cite{xie2021segformer} trained on Cityscapes~\cite{cordts2016cityscapes}) are discarded, removing 22\% of image-mask pairs. During training and inference, up to 8 street-level images are sampled per segment (\cref{fig:sl-stats,fig:sl-challenges,fig:sl-occlusion}).

\paragraph{Splits.} The dataset is split 85\%/7.5\%/7.5\% at the segment level into train, validation, and test sets.

\begin{figure}[t]
  \vspace{4mm}
  \centering
  \includegraphics[width=\linewidth]{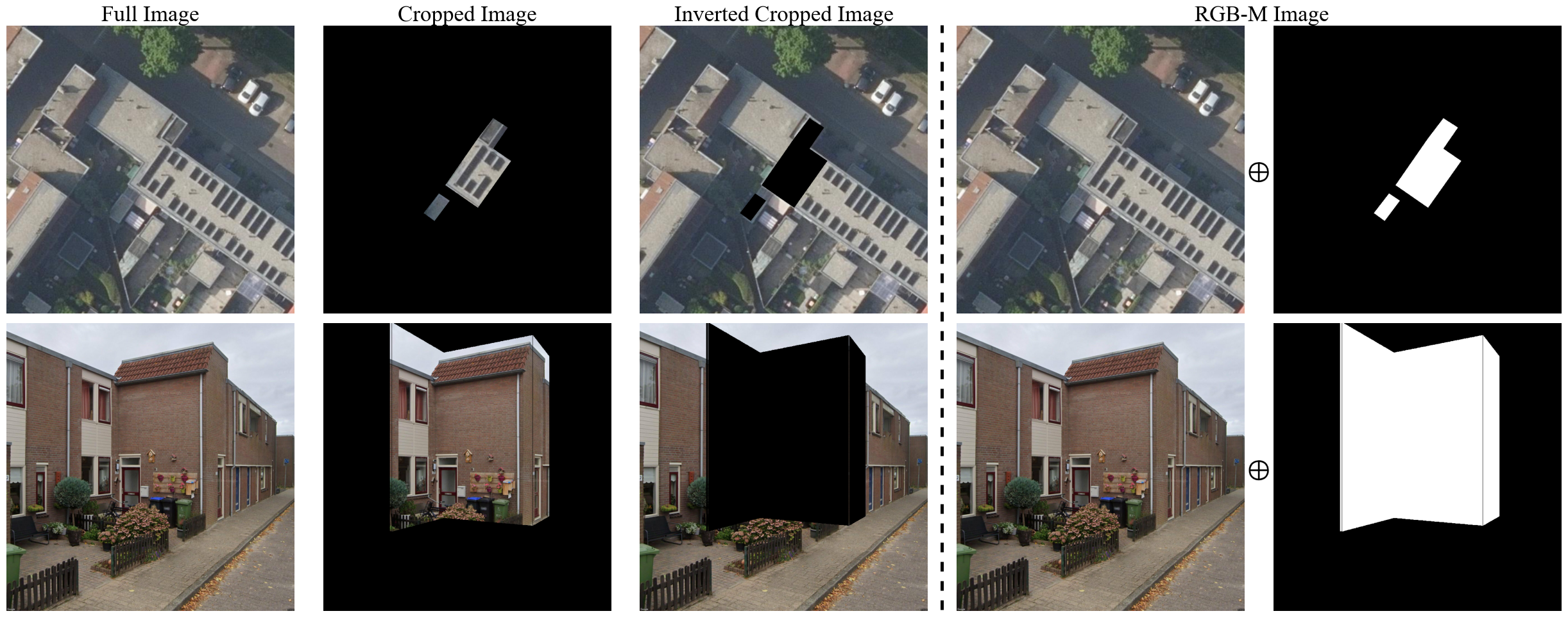}
  \vspace{2mm}
  \caption{Four masking strategies applied to satellite (top) and street-level (bottom) imagery. From left to right: full image, cropped (background zeroed), inverted crop (building zeroed), and RGB-M (mask appended as 4th channel).}
  \label{fig:masking}
\end{figure}

\newpage
\subsection{Masking Strategies}
\label{sec:masking}

To isolate the target building in each image, we evaluate four masking strategies (\cref{fig:masking}) using the binary building footprint mask $M$:

\begin{enumerate}
    \item \textbf{Full image}: The unmodified RGB image ($\mathbb{R}^{H \times W \times 3}$).
    \item \textbf{Crop}: Element-wise product $I_\text{rgb} \odot M$, zeroing out all pixels outside the building footprint.
    \item \textbf{Inverted crop}: $I_\text{rgb} \odot (1 - M)$, retaining only the surrounding context as a sanity check.
    \item \textbf{RGB-M}: The mask is appended as a fourth input channel, $[I_\text{rgb}; M] \in \mathbb{R}^{H \times W \times 4}$, letting the model learn to balance building focus with contextual information.
\end{enumerate}

\begin{figure}[t]
  \vspace{4mm}
  \centering
  \includegraphics[width=\linewidth]{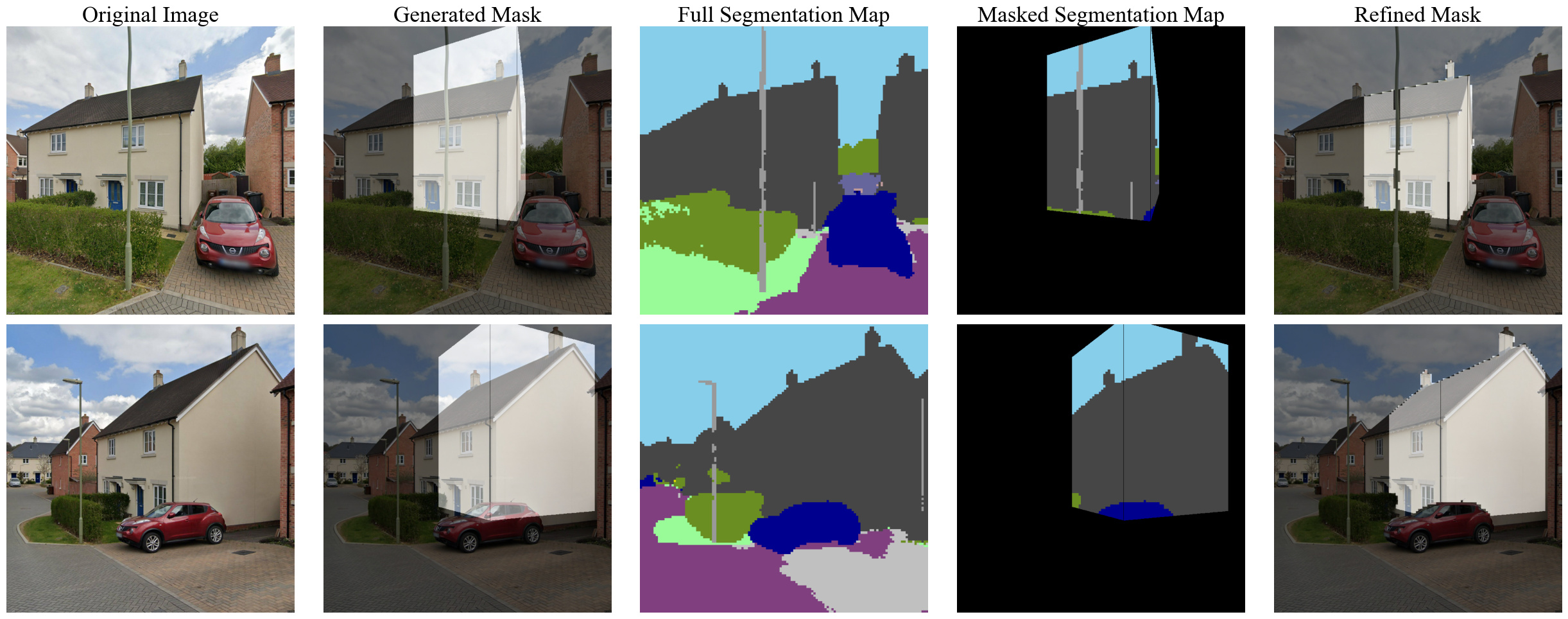}
  \vspace{2mm}
  \caption{Street-level mask generation pipeline. From left to right: original image, projected building footprint mask, full semantic segmentation map (SegFormer-B0), masked segmentation, and refined mask.}
  \label{fig:sl-pipeline}
  \vspace{4mm}
\end{figure}

\begin{figure}[t]
  \vspace{4mm}
  \centering
  \includegraphics[width=\linewidth]{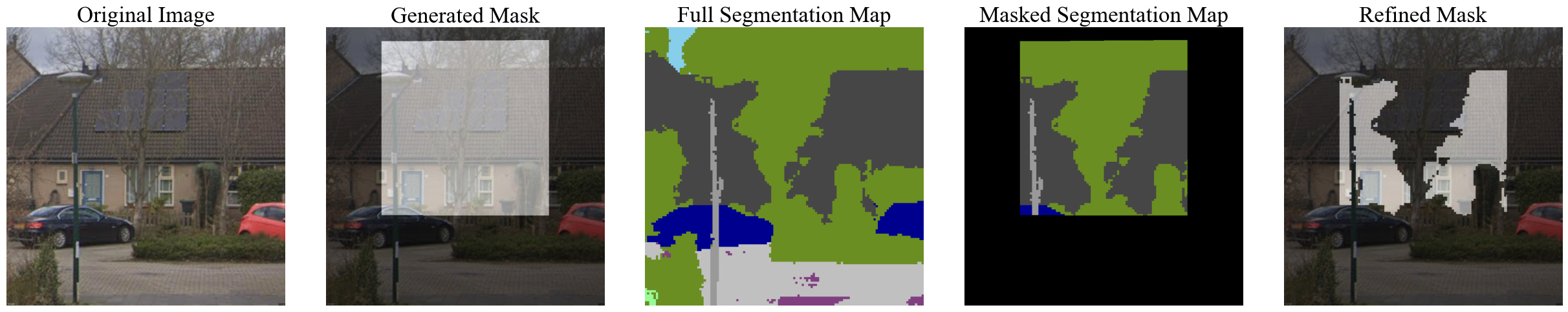}
  \caption{Mask refinement limitation: sparse vegetation in the segmentation map causes relevant building areas to be incorrectly excluded from the refined mask.}
  \label{fig:mask-refinement}
\end{figure}

\vspace{2mm}
For street-level imagery, pixel-level masks are generated by projecting the building footprint into the camera view. \Cref{fig:sl-pipeline} illustrates the mask generation pipeline, and \cref{fig:mask-refinement} shows a limitation of the refinement step where sparse vegetation causes relevant building areas to be excluded.

\vspace{3mm}
\subsection{Uni-modal Models}
\label{sec:unimodal}

\paragraph{Backbone.} We compare ResNet-50~\cite{he2016deep} and DINOv2-S~\cite{oquab2023dinov2}. For ResNet-50, a global average-pooled feature vector serves as the building representation. For DINOv2-S, we use the \texttt{[CLS]} token output. Both feed into task-specific linear classifier heads for roof elements and roof materials. For the RGB-M masking strategy, the first convolutional layer (ResNet-50) or patch embedding projection (DINOv2-S) is adapted to accept 4-channel input.

\begin{figure}[t]
  \centering
  \hspace{-0.2\linewidth}\includegraphics[width=1.2\linewidth]{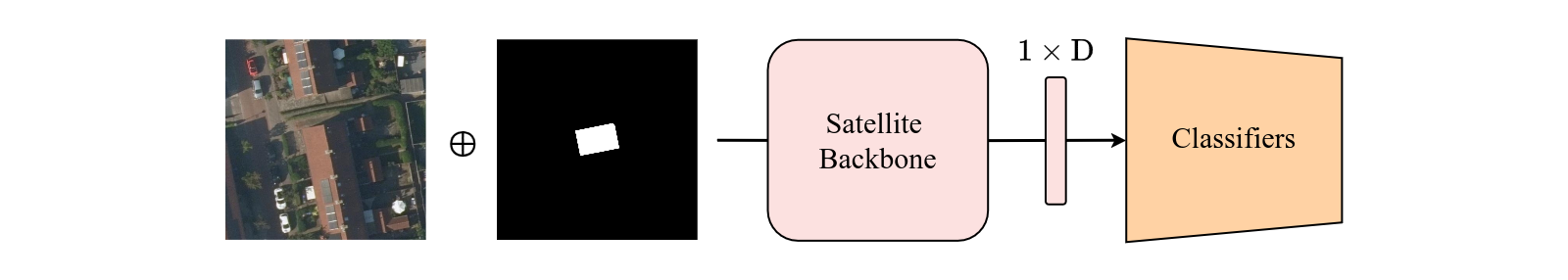}
  \caption{Uni-modal satellite model. The RGB-M input (4 channels) is encoded by the backbone into a $1 \times D$ feature vector and passed to task-specific classifiers.}
  \label{fig:uni-sat}
\end{figure}

\paragraph{Satellite model.} A single satellite image is encoded into a global feature vector and passed to both classifier heads (\cref{fig:uni-sat}). Since each building has exactly one satellite view, no aggregation is needed; the backbone output directly serves as the building-level representation.

\begin{figure}[t]
  \centering
  \hspace{-0.02\linewidth}\includegraphics[width=1.02\linewidth]{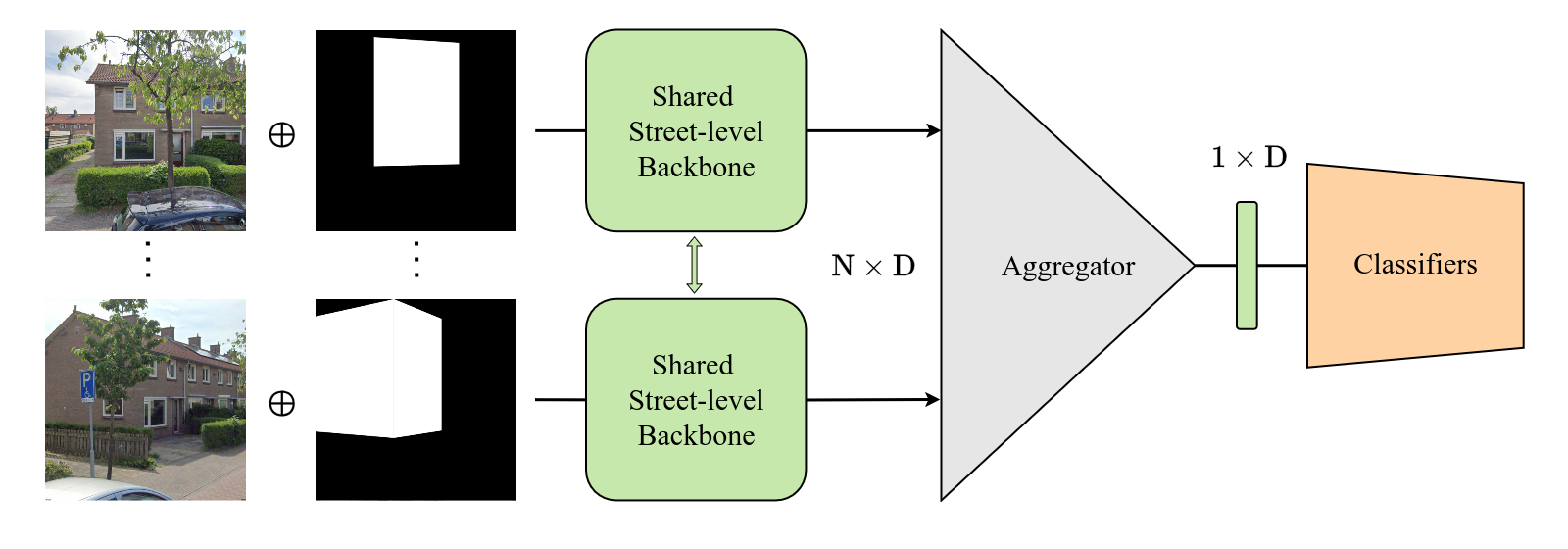}
  \caption{Uni-modal street-level model. Each street-level image is independently encoded by a shared backbone. Feature vectors are aggregated (max/mean/attention pooling) into a single representation for classification.}
  \label{fig:uni-sl}
\end{figure}

\paragraph{Street-level model.} Each street-level image is independently encoded by the same backbone (\cref{fig:uni-sl}). The resulting feature vectors are aggregated via max pooling across views and passed to the classifiers. We also evaluate mean pooling and attention-based pooling~\cite{ilse2018attention}; max pooling performs best overall.

\subsection{Multi-modal Fusion}
\label{sec:fusion}

We evaluate three fusion strategies of increasing complexity, all using DINOv2-S with RGB-M masking.

\begin{figure}[t]
  \centering
  \hspace{-0.02\linewidth}\includegraphics[width=1.02\linewidth]{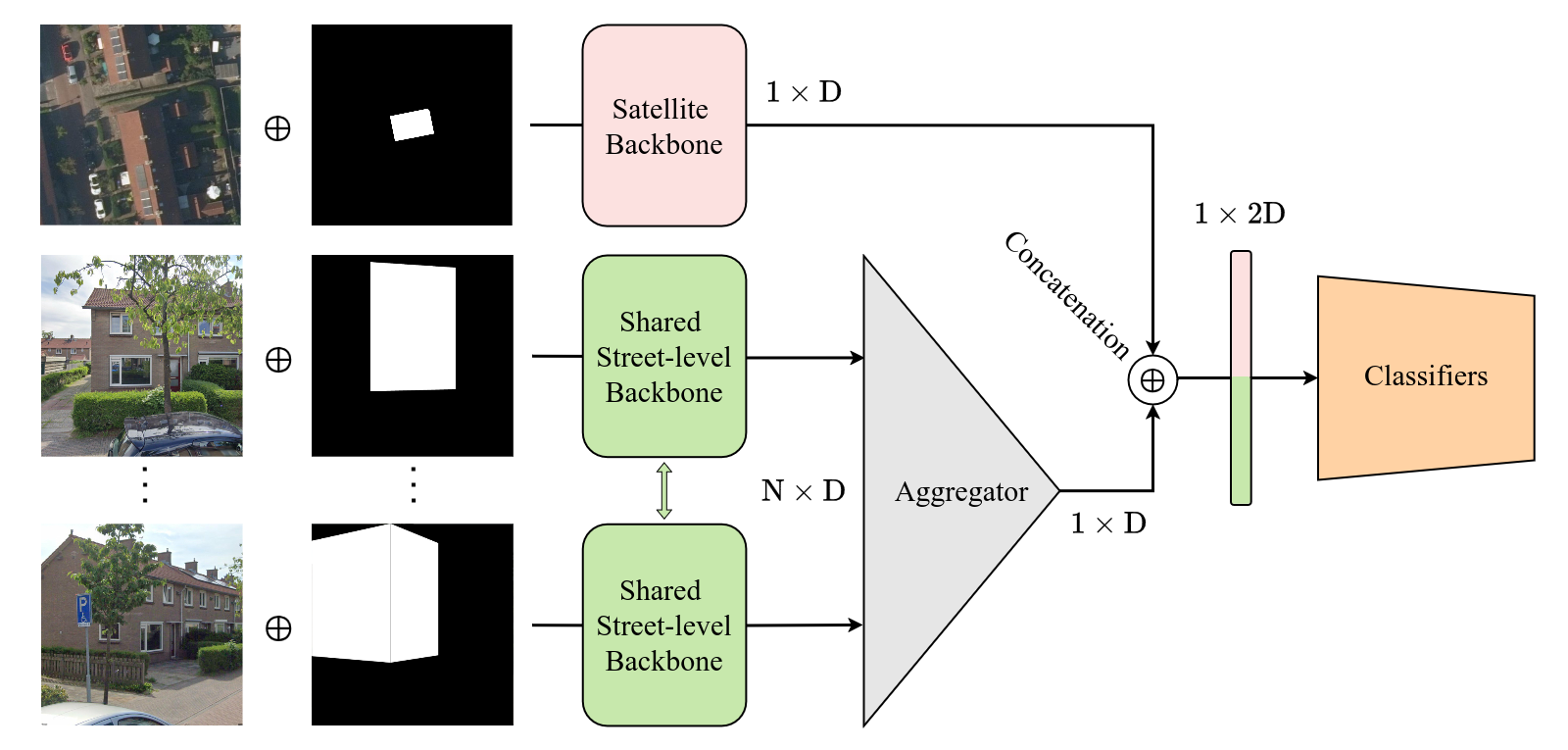}
  \caption{Concatenation fusion baseline. Satellite and aggregated street-level feature vectors are concatenated and passed to classifiers. Zero additional fusion parameters.}
  \label{fig:fusion-concat}
\end{figure}

\paragraph{Concatenation.} The satellite feature vector and the max-pooled street-level feature vector are concatenated and passed to the classifiers (\cref{fig:fusion-concat}). When no street-level images are available, a learnable placeholder vector is substituted. This baseline has zero additional fusion parameters.

\begin{figure}[t]
  \centering
  \hspace{-0.02\linewidth}\includegraphics[width=1.02\linewidth]{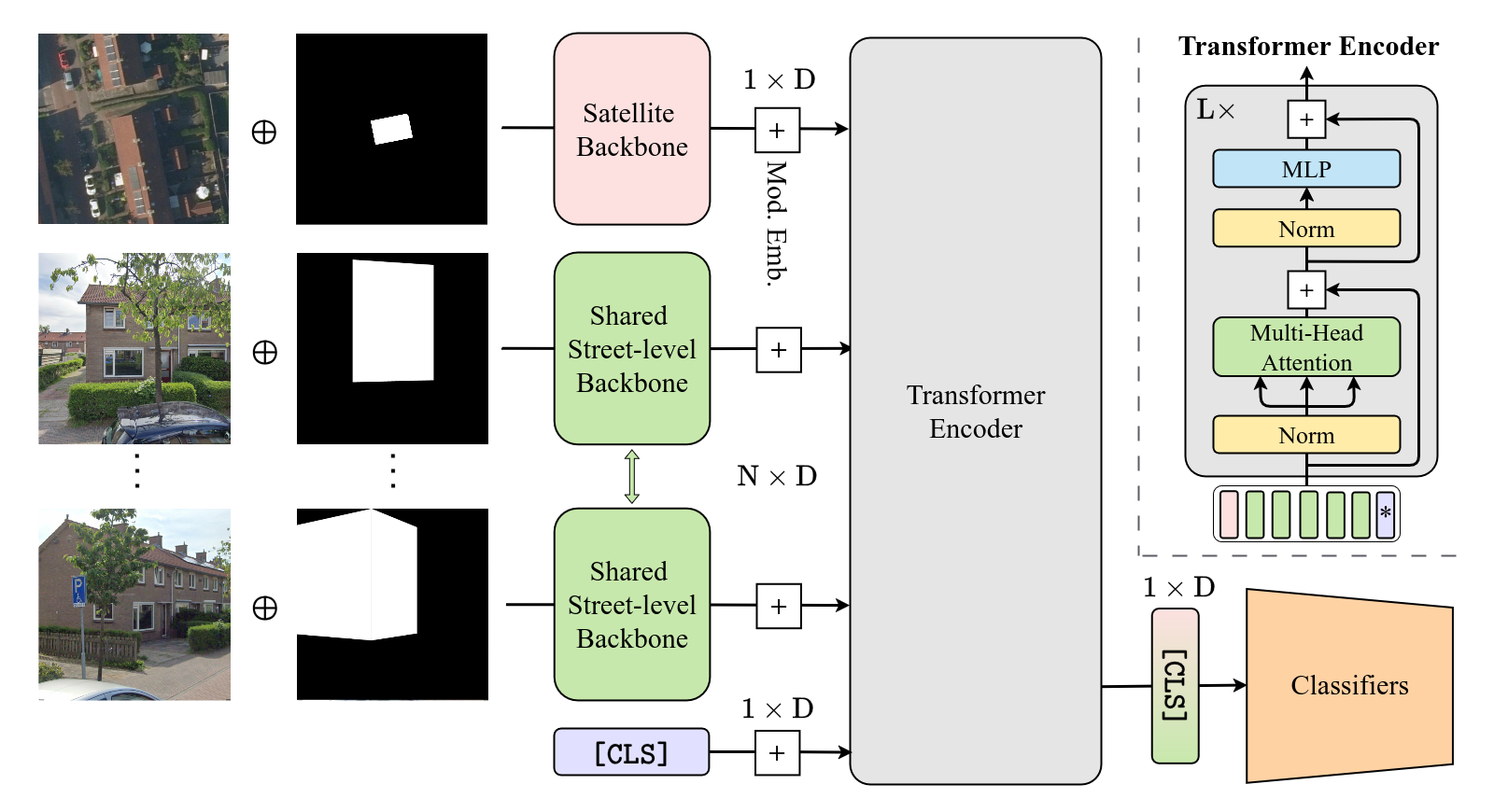}
  \caption{Feature Vector Transformer fusion. Individual feature vectors from all views are combined with a learnable \texttt{[CLS]} token and modality embeddings, then processed by a transformer encoder.}
  \label{fig:fusion-fvt}
\end{figure}

\paragraph{Feature Vector Transformer.} Feature vectors from the satellite image and each individual street-level view are combined into a single sequence (\cref{fig:fusion-fvt}), augmented with a learnable \texttt{[CLS]} token and modality embeddings. A standard transformer encoder with $L$ layers and $h{=}8$ attention heads processes this sequence; the \texttt{[CLS]} output is used for classification. This eliminates the need for explicit pooling and handles variable input counts.

\begin{figure*}[t]
  \centering
  \includegraphics[width=0.8\linewidth]{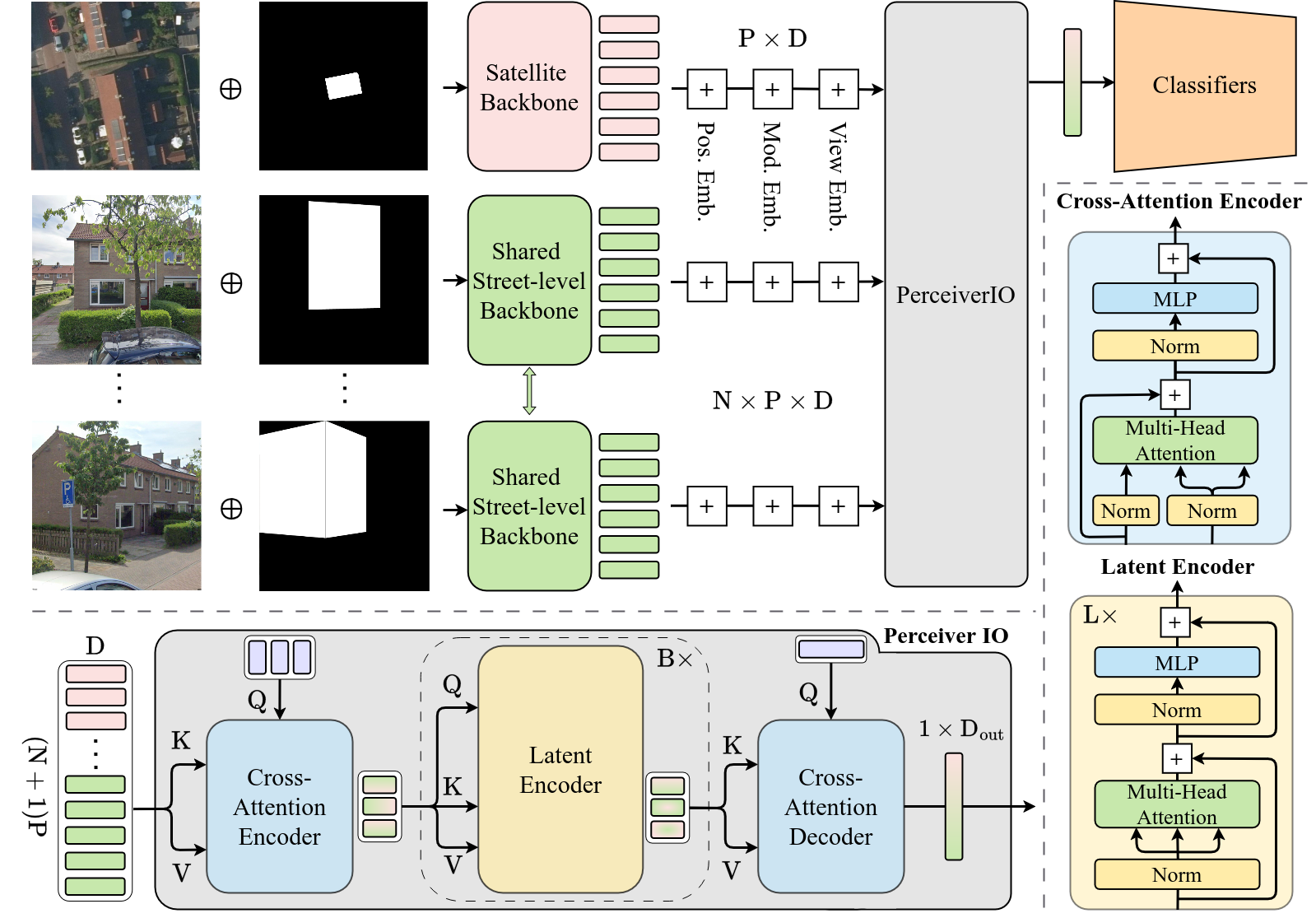}
  \caption{Perceiver IO fusion architecture. A satellite image and $N$ street-level images are encoded by shared DINOv2 backbones into spatial patch tokens, augmented with positional, modality, and view embeddings. The cross-attention encoder compresses the variable-length token sequence into a fixed-size latent array. After refinement through $B$ blocks of self-attention layers, a cross-attention decoder produces a global embedding for classification.}
  \label{fig:perceiver}
\end{figure*}

\newpage
\paragraph{Perceiver IO.} Our main contribution (\cref{fig:perceiver}) operates on \emph{spatial tokens} rather than pooled feature vectors, preserving fine-grained patch-level detail. Given a satellite image and $N$ street-level images, each processed by the DINOv2-S backbone into $P$ patch tokens of dimension $D$, the full input sequence is $X \in \mathbb{R}^{(N+1)P \times D}$. Each token is augmented with three learned embeddings: (i) a 2D positional embedding encoding its spatial location within the source image, (ii) a modality embedding distinguishing satellite from street-level tokens, and (iii) a view embedding for inter-view localization.

The Perceiver IO processes this variable-length input in three stages:
\begin{enumerate}
    \item \textbf{Cross-attention encoder.} A single-head cross-attention layer compresses $X$ into a fixed-size latent array $Z_0 \in \mathbb{R}^{N_z \times D_z}$ using a learnable query array, where $N_z \ll (N{+}1)P$.
    \item \textbf{Latent refinement.} $B$ blocks of $L$ self-attention sub-layers (with parameters shared across blocks but not within) refine the latent representation.
    \item \textbf{Cross-attention decoder.} A single-head cross-attention layer with a learnable output query $q \in \mathbb{R}^{1 \times D_z}$ attends over the refined latents to produce a global embedding $y \in \mathbb{R}^{D_\text{out}}$, which is passed to the classifier heads.
\end{enumerate}

This architecture compresses an arbitrarily long token sequence into a fixed-size bottleneck, naturally handling any number of street-level views without padding or truncation.

\subsection{Training}
\label{sec:training}

The total loss is $\mathcal{L} = 0.5 \cdot \mathcal{L}_\text{RE} + 0.5 \cdot \mathcal{L}_\text{RM}$, where $\mathcal{L}_\text{RE}$ and $\mathcal{L}_\text{RM}$ are binary cross-entropy losses for roof elements and roof materials respectively. We use AdamW with weight decay 0.05, separate learning rates for the backbone ($5 \times 10^{-5}$) and fusion heads ($5 \times 10^{-4}$), linear warm-up (5 iterations) followed by cosine annealing ($T_\text{max}{=}50$), and early stopping with patience~8. Data augmentation includes random horizontal flips, rotations (up to $180^\circ$ for satellite, $8^\circ$ for street-level), brightness/contrast jitter, color jitter, and Gaussian blur for satellite images; street-level images receive a lighter subset given their natural variability. Images are resized to $518 \times 518$ pixels (DINOv2-S's native resolution). Training uses mixed-precision (float16) on 2$\times$ NVIDIA RTX 3090 GPUs with batch size 32 for the Perceiver IO model. A custom dataloader buckets samples by street-level image count, avoiding padding overhead.

\section{Experiments}
\label{sec:experiments}

We evaluate along four axes: (1)~masking strategies, (2)~backbone and fine-tuning, (3)~input resolution, and (4)~fusion architecture. All results are reported as macro-averaged mean Average Precision (mAP) over classes. For roof materials, we additionally report mAP$^*$, computed over a reliable subset of five classes that excludes thatch and glass, which have very few training samples (184 and 280 respectively).

\subsection{Masking Strategies}
\label{sec:exp-masking}

\begin{table}[t]
\centering
\caption{Satellite masking strategies. Best per backbone in \textbf{bold}.}
\label{tab:masking-sat}
\small
\begin{tabular}{llccc}
\toprule
Backbone & Masking & Elem & Mat & Mat$^*$ \\
\midrule
\multirow{4}{*}{ResNet-50}
& Full     & 0.867 & 0.694 & 0.754 \\
& Crop     & \textbf{0.928} & 0.701 & 0.743 \\
& Inv-Crop & 0.650 & 0.502 & 0.649 \\
& RGB-M    & 0.924 & \textbf{0.721} & \textbf{0.769} \\
\midrule
\multirow{4}{*}{DINOv2-S}
& Full     & 0.891 & \textbf{0.736} & \textbf{0.788} \\
& Crop     & 0.927 & 0.688 & 0.733 \\
& Inv-Crop & 0.696 & 0.621 & 0.724 \\
& RGB-M    & \textbf{0.939} & 0.729 & \textbf{0.788} \\
\bottomrule
\end{tabular}
\end{table}

\begin{table}[t]
\centering
\caption{Street-level masking strategies. Best in \textbf{bold}.}
\label{tab:masking-sl}
\small
\begin{tabular}{llccc}
\toprule
Backbone & Masking & Elem & Mat & Mat$^*$ \\
\midrule
\multirow{4}{*}{ResNet-50}
& Full     & 0.596 & 0.526 & 0.613 \\
& Crop     & \textbf{0.684} & \textbf{0.554} & \textbf{0.660} \\
& Inv-Crop & 0.595 & 0.507 & 0.608 \\
& RGB-M    & 0.626 & 0.546 & 0.639 \\
\midrule
\multirow{4}{*}{DINOv2-S}
& Full     & 0.635 & 0.551 & 0.648 \\
& Crop     & 0.666 & 0.558 & 0.666 \\
& Inv-Crop & 0.639 & 0.563 & 0.644 \\
& RGB-M    & \textbf{0.712} & \textbf{0.597} & \textbf{0.687} \\
\bottomrule
\end{tabular}
\end{table}

\Cref{tab:masking-sat,tab:masking-sl} compare the four masking strategies on satellite and street-level imagery, respectively. Across both modalities, DINOv2-S with RGB-M masking achieves the best overall performance: 0.939 mAP for roof elements and 0.729 for roof materials on satellite, 0.712 and 0.597 on street-level. The RGB-M strategy consistently outperforms alternatives for DINOv2-S by providing the building footprint as a soft spatial prior without discarding contextual information. Hard cropping is competitive for ResNet-50 on satellite (0.928 mAP for roof elements) but degrades more for DINOv2-S, likely because the zero-padded background conflicts with DINOv2's pretrained patch statistics. The inverted-crop sanity check confirms that surrounding context alone carries meaningful signal (0.696 mAP for roof elements on DINOv2-S satellite), but performance is substantially lower than mask-inclusive strategies.

\begin{figure}[t]
  \centering
  \includegraphics[width=\linewidth]{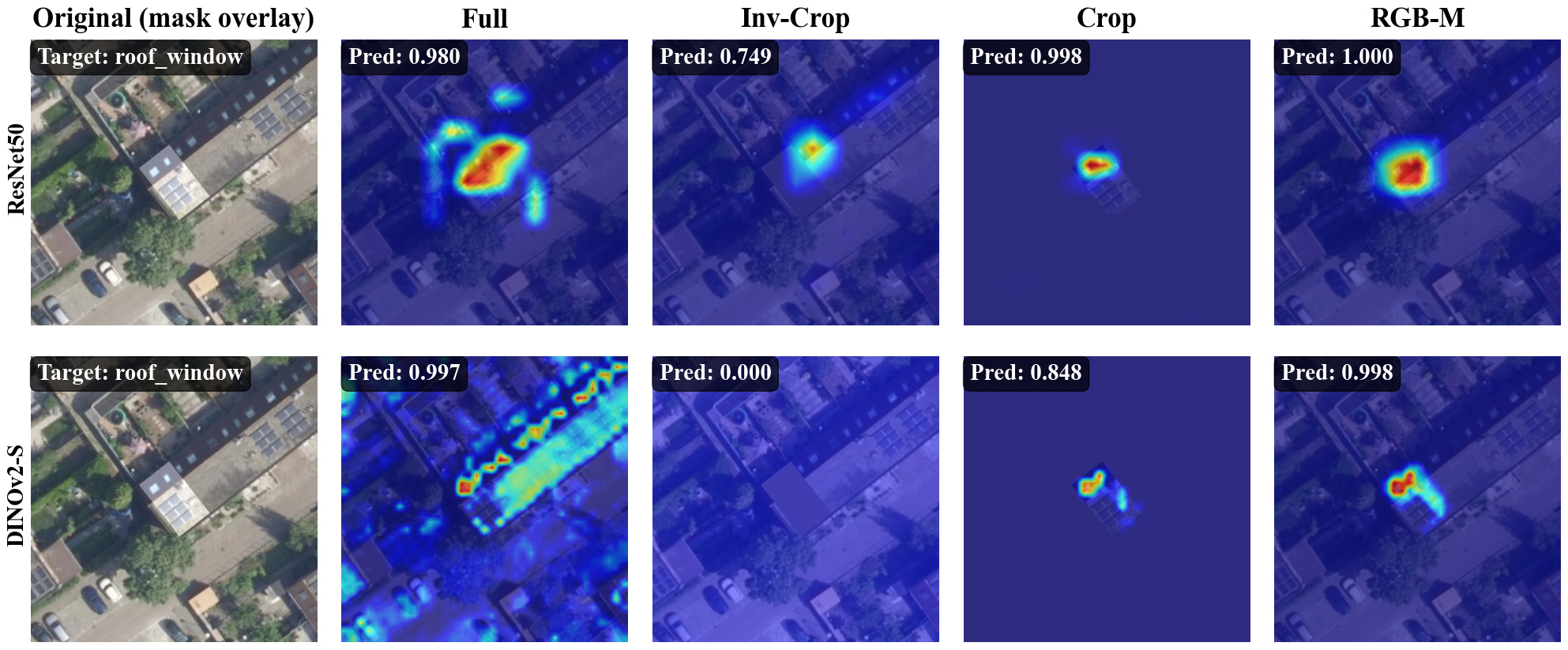}
  \caption{Grad-CAM visualizations for satellite masking strategies (ResNet-50 top, DINOv2-S bottom). Full image attention is diffuse; Crop forces focus onto the building; Inv-Crop attends to context only.}
  \label{fig:gradcam-sat}
\end{figure}

\begin{figure}[t]
  \centering
  \includegraphics[width=\linewidth]{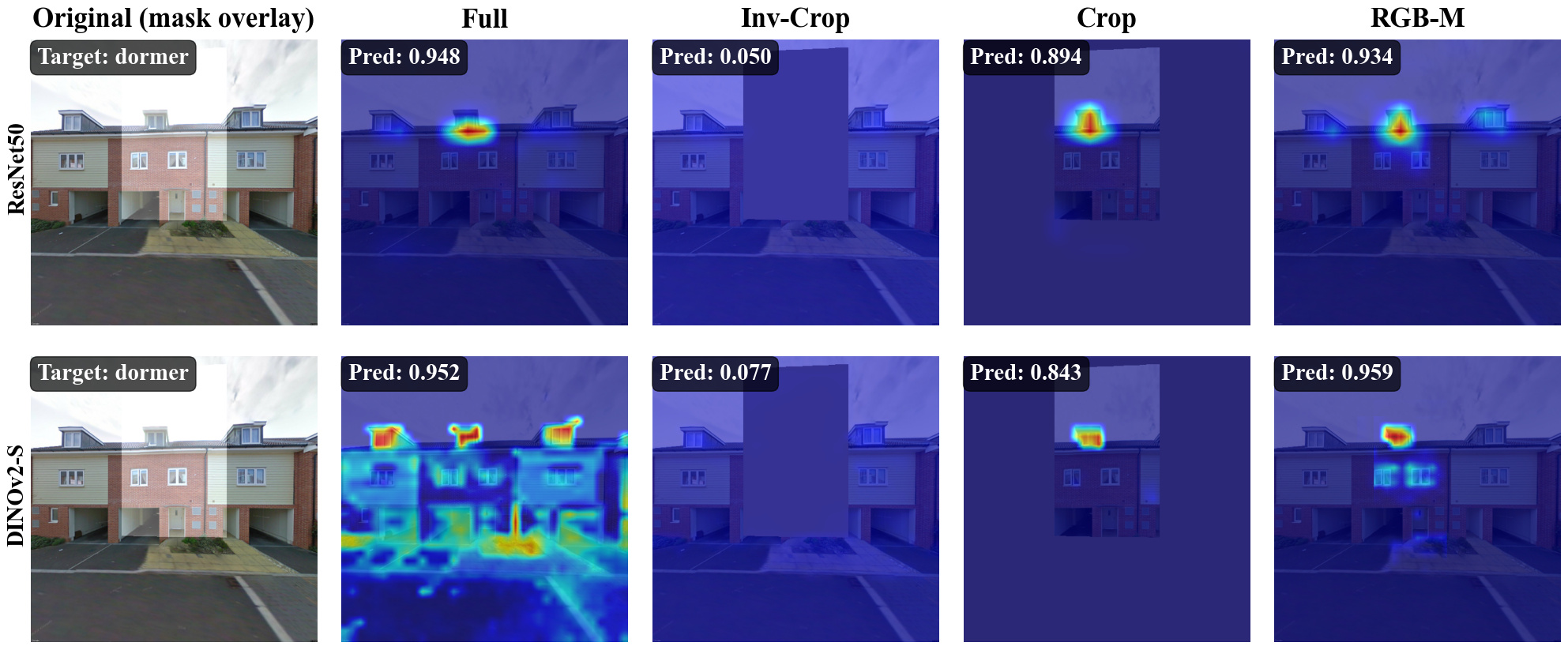}
  \caption{Grad-CAM visualizations for street-level masking strategies (ResNet-50 top, DINOv2-S bottom). Without masking, attention is spread; Crop isolates the building facade.}
  \label{fig:gradcam-sl}
\end{figure}

Grad-CAM visualizations confirm these findings qualitatively (\cref{fig:gradcam-sat,fig:gradcam-sl}). Without masking, attention is diffusely spread across the entire scene. Hard cropping forces the model to focus exclusively on the building footprint, while the inverted crop attends only to context. RGB-M produces the most balanced activation pattern: the model attends primarily to the building while retaining access to contextual cues at the boundary, consistent with its superior quantitative performance.

We also compared the standard footprint-projected masks against refined masks produced by the SegFormer-B0 segmentation model, with and without the 20\% visibility threshold. The standard mask consistently outperforms the refined version, suggesting that segmentation-based refinement can remove useful building regions (e.g., behind sparse vegetation) without compensating gains. Applying the visibility threshold has little effect with standard masks, indicating that the model is reasonably robust to partially occluded views.

\subsection{Backbone and Fine-tuning}
\label{sec:exp-finetuning}

\Cref{tab:finetuning} shows that full fine-tuning of DINOv2-S yields the highest mAP for both tasks, with a large gap over linear probing (+22.4 mAP for roof elements, +18.1 mAP for roof materials). Partial strategies (last-3 layers, progressive unfreezing) close most of the gap but fall short of full adaptation, suggesting the domain shift from natural images to aerial/street-level building imagery requires updating early feature layers. The same pattern holds for street-level imagery, where full fine-tuning achieves 0.703 mAP for roof elements vs.\ 0.591 for linear probing.

\subsection{Input Resolution}
\label{sec:exp-resolution}

Performance peaks at $518 \times 518$ (\cref{tab:resolution}), which matches DINOv2-S's pretraining resolution. Higher resolution (784) does not help, likely because the interpolated positional embeddings introduce artifacts. In contrast, ResNet-50 benefits from higher resolution up to 784, consistent with its lack of resolution-dependent positional priors.

\begin{table}[t]
\centering
\caption{DINOv2-S fine-tuning strategies on satellite imagery with RGB-M masking.}
\label{tab:finetuning}
\small
\begin{tabular}{lccc}
\toprule
Strategy & Elem & Mat & Mat$^*$ \\
\midrule
Linear probing   & 0.715 & 0.548 & 0.651 \\
Last-3 layers     & 0.912 & 0.724 & 0.784 \\
Progressive       & 0.912 & 0.725 & 0.792 \\
Full              & \textbf{0.939} & \textbf{0.729} & 0.786 \\
\bottomrule
\end{tabular}
\end{table}

\subsection{Fusion Architectures}
\label{sec:exp-fusion}

\begin{figure}[t]
  \centering
  \includegraphics[width=\linewidth]{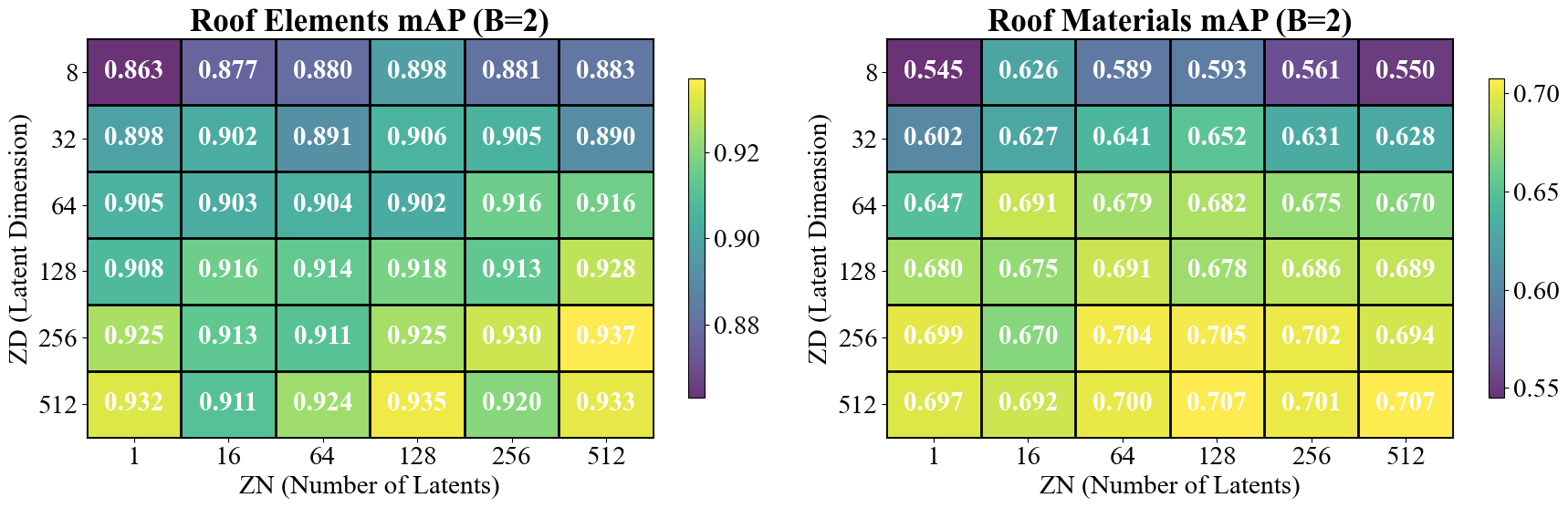}
  \caption{Perceiver IO sensitivity to latent configuration ($B{=}2$). mAP for roof elements (left) and materials (right) as a function of the number of latents $N_z$ and latent dimensionality $D_z$. Performance depends primarily on $D_z$; even $N_z{=}1$ suffices at high dimensionality.}
  \label{fig:heatmap}
\end{figure}

\paragraph{Perceiver IO configuration.} \Cref{fig:heatmap} shows the sensitivity to latent configuration. We sweep over the number of latent vectors $N_z \in \{1, 16, 64, 128, 256, 512\}$, latent dimensionality $D_z \in \{8, 32, 64, 128, 256, 512\}$, and refinement blocks $B \in \{0, 1, 2, 4\}$. Latent dimensionality has a much stronger impact than the number of latents; even $N_z{=}1$ achieves near-peak performance with $D_z{=}512$. We select $N_z{=}128$, $D_z{=}512$, $B{=}2$ for final experiments as the best trade-off.

\begin{table}[t]
\centering
\caption{Impact of input resolution on DINOv2-S satellite performance (RGB-M masking).}
\vspace{-1mm}
\label{tab:resolution}
\small
\begin{tabular}{lcc}
\toprule
Resolution & Elem & Mat \\
\midrule
$224 \times 224$ & 0.904 & 0.710 \\
$336 \times 336$ & 0.924 & 0.711 \\
$518 \times 518$ & \textbf{0.939} & \textbf{0.729} \\
$784 \times 784$ & 0.928 & 0.721 \\
\bottomrule
\end{tabular}
\vspace{-2mm}
\end{table}

\begin{table}[t]
\centering
\caption{Overall model comparison. All models use DINOv2-S with RGB-M masking. Best overall in \textbf{bold}; best fusion model \underline{underlined}.}
\vspace{-1mm}
\label{tab:overall}
\small
\begin{tabular}{lrccc}
\toprule
Model & Params & Elem & Mat & Mat$^*$ \\
\midrule
Satellite only     & --      & \textbf{0.939} & \textbf{0.729} & \textbf{0.786} \\
Street-level only  & --      & 0.712 & 0.597 & 0.687 \\
\midrule
Concatenation      & 0M      & 0.912 & 0.714 & 0.782 \\
FV Transformer     & 13.0M   & 0.927 & 0.705 & 0.767 \\
Perceiver IO       & 25.7M   & \underline{0.935} & \underline{0.707} & \underline{0.774} \\
\bottomrule
\end{tabular}
\vspace{-2mm}
\end{table}

\begin{table*}[t]
\centering
\caption{Per-class Average Precision for the satellite-only baseline and Perceiver IO fusion model. Classes where fusion improves by $\geq$1 AP point are highlighted in \textbf{bold}.}
\label{tab:perclass}
\small
\begin{tabular}{lcccccc|ccccccc}
\toprule
& \multicolumn{6}{c|}{Roof Elements} & \multicolumn{7}{c}{Roof Materials} \\
\cmidrule(lr){2-7} \cmidrule(lr){8-14}
Model & Solar & Dorm & Sky & Win & Chim & Ext & Bit & Slate & Tiles & Alu & Thatch & Corr & Glass \\
\midrule
Satellite only & 0.979 & 0.932 & \textbf{0.882} & 0.971 & 0.976 & \textbf{0.892} & \textbf{0.961} & 0.778 & 0.946 & \textbf{0.637} & \textbf{0.876} & \textbf{0.609} & 0.294 \\
Perceiver IO   & 0.981 & \textbf{0.945} & 0.858 & \textbf{0.980} & 0.978 & 0.867 & 0.944 & \textbf{0.891} & \textbf{0.956} & 0.591 & 0.741 & 0.490 & \textbf{0.339} \\
\midrule
$\Delta$ & +0.2 & +1.3 & $-$2.4 & +0.9 & +0.2 & $-$2.5 & $-$1.7 & +11.3 & +1.0 & $-$4.6 & $-$13.5 & $-$11.9 & +4.5 \\
\bottomrule
\end{tabular}
\end{table*}

\Cref{tab:overall} presents the overall comparison. The satellite-only model achieves the highest macro-averaged mAP, with Perceiver IO as the best fusion method (0.935 for elements, 0.707 for materials). However, macro mAP obscures important per-class differences. \Cref{tab:perclass} reveals that the Perceiver IO model substantially improves on classes where street-level imagery provides complementary information: slate gains +11.3 AP (0.778 $\to$ 0.891), dormer +1.3 AP, tiles +1.0 AP, roof window +0.9 AP, and glass +4.5 AP. In contrast, the satellite-only model retains an advantage for classes that are primarily visible from above: skylight, external installations, bitumen, aluminium, thatch, and corrugated sheets. For these classes, street-level views introduce noise rather than useful signal, as the relevant features are occluded or absent at ground level.

The concatenation baseline (0.912 mAP for roof elements) underperforms both the satellite-only model and the Perceiver IO, indicating that naive feature-level fusion degrades the satellite signal. The Feature Vector Transformer (0.927 mAP for roof elements) improves over concatenation but still loses spatial detail through global pooling before fusion; sweeping the number of encoder layers ($L \in \{1,2,4,6\}$) shows that $L{=}1$ performs comparably to deeper configurations, so the additional capacity provides little benefit at the feature-vector level. Perceiver IO's token-level fusion recovers much of this lost information.

\section{Discussion}
\label{sec:discussion}

\paragraph{When does fusion help?}
The per-class analysis in \cref{tab:perclass} reveals a clear pattern: fusion improves classification for attributes that are visible and distinctive in street-level imagery. Slate and tiles, for example, are visually similar from above but readily distinguishable at ground level through texture and color differences, explaining the +11.3 AP gain for slate. Dormers and roof windows protrude from the roof plane and are often visible from the street, yielding consistent improvements. In contrast, skylights, external installations, and bitumen roofing are best observed from above, and adding street-level tokens for these classes dilutes the satellite signal.

\begin{figure}[t]
  \centering
  \includegraphics[width=\linewidth]{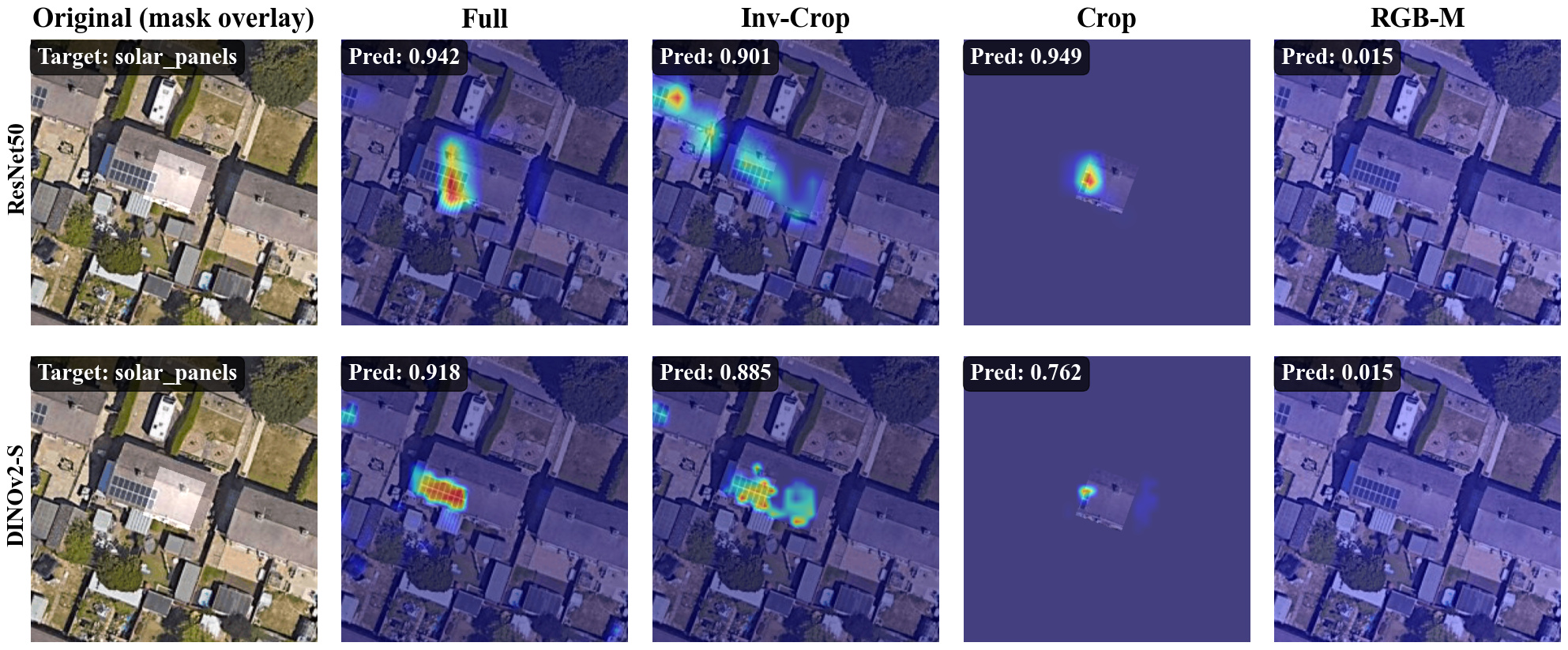}
  \caption{Grad-CAM visualizations for satellite imagery across masking strategies (ResNet-50 top, DINOv2-S bottom). RGB-M focuses attention on the target building while retaining contextual awareness. Hard cropping forces attention onto the masked region but removes useful context.}
  \label{fig:gradcam}
\end{figure}

\begin{figure}[t]
  \centering
  \includegraphics[width=\linewidth]{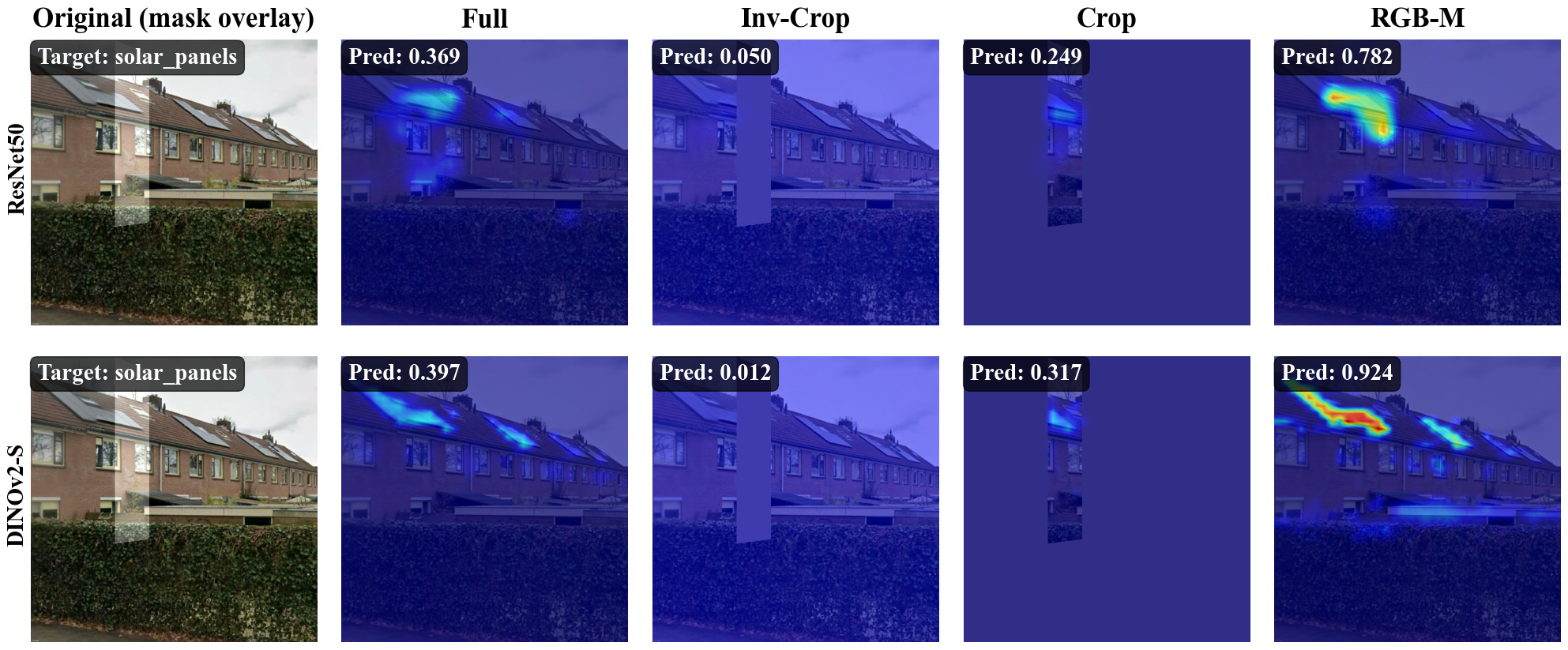}
  \caption{Grad-CAM visualizations for street-level imagery with RGB-M masking (ResNet-50 top, DINOv2-S bottom). RGB-M effectively focuses attention on the target building facade even in cluttered street scenes.}
  \label{fig:gradcam-sl-rgbm}
\end{figure}

\paragraph{RGB-M as a soft spatial prior.}
The RGB-M masking strategy outperforms hard cropping because it provides the building footprint as an additional input channel rather than destructively zeroing out context pixels. This is especially important for DINOv2-S, whose pretrained patch statistics assume complete image patches. Hard cropping introduces zero-filled patches that fall outside the pretraining distribution, degrading feature quality. The inverted-crop experiment confirms that context alone carries meaningful signal---nearby buildings, street patterns, and vegetation provide cues about the target building's type and materials. \Cref{fig:gradcam,fig:gradcam-sl-rgbm} illustrate this effect for both satellite and street-level imagery.

\paragraph{DINOv2 vs.\ ResNet-50.}
Fully fine-tuned DINOv2-S consistently outperforms ResNet-50 despite having a similar parameter count. The gap is largest for roof materials (+0.8 mAP on satellite), suggesting that DINOv2's self-supervised pretraining produces features that better capture fine-grained material textures. Linear probing performs poorly (0.715 mAP for roof elements), indicating that the domain gap between natural images and building inspection imagery requires substantial feature adaptation.

\begin{figure*}[t]
  \centering
  \includegraphics[width=0.9\linewidth]{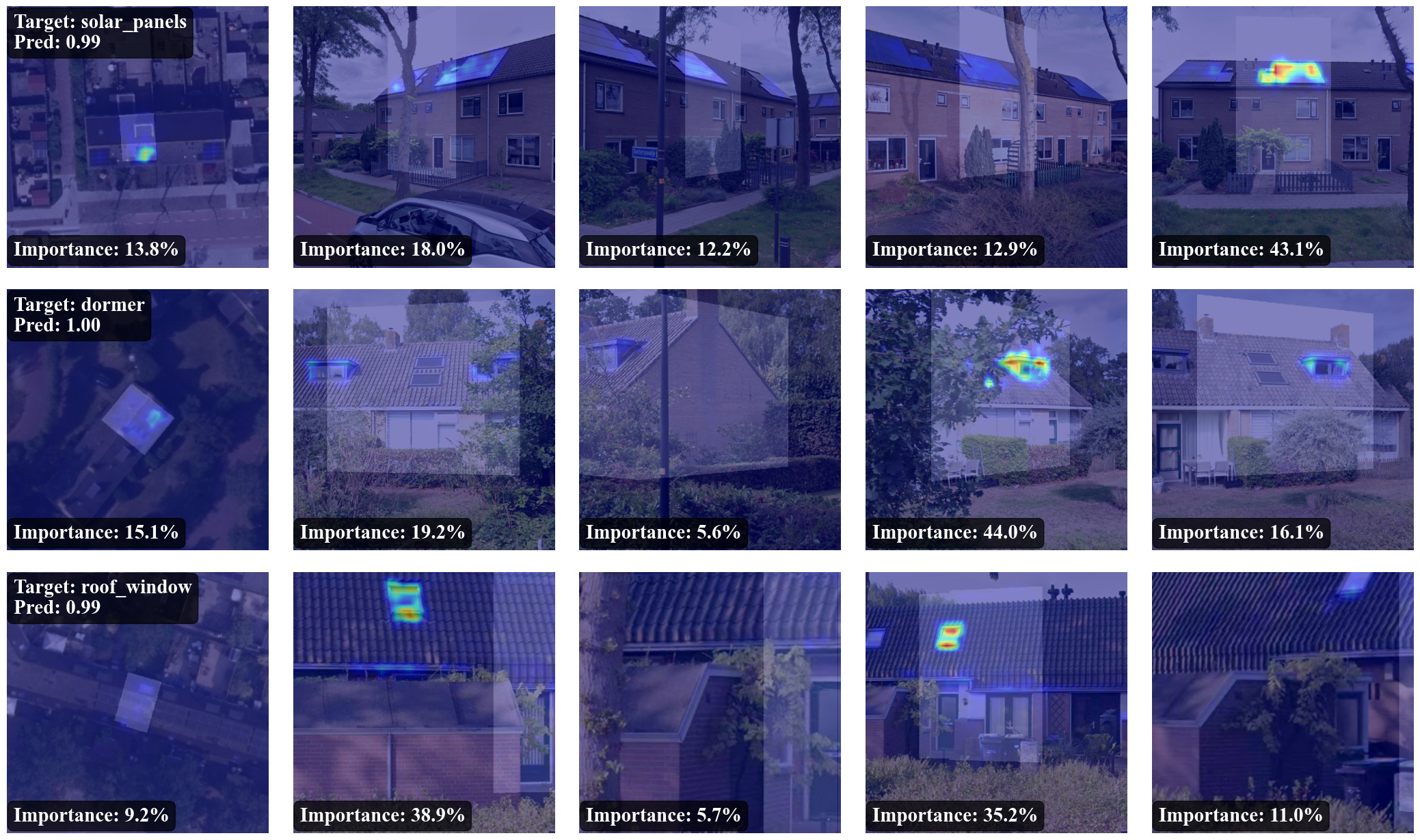}
  \caption{Perceiver IO attention rollout visualizations for three buildings. Each row shows the satellite image (left, with attention overlay) and street-level views with per-view importance scores. The model correctly localizes solar panels (top), dormers (middle), and roof windows (bottom) across modalities, assigning higher importance to views where the target attribute is visible.}
  \label{fig:attention}
\end{figure*}

\begin{figure}[t]
  \centering
  \includegraphics[width=\linewidth]{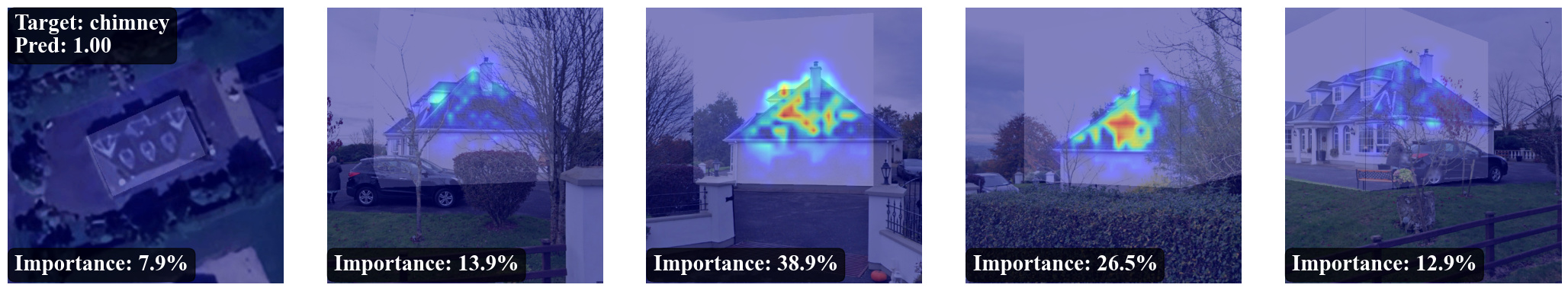}
  \caption{Perceiver IO attention rollout for chimney detection---a failure case. Attention is diffusely spread across views without clear localization, suggesting the model may rely on contextual shortcuts rather than localizing the chimney directly.}
  \label{fig:attention-bad}
\end{figure}

\paragraph{Perceiver IO design choices.}
Latent dimensionality ($D_z$) matters more than the number of latents ($N_z$): reducing $D_z$ from 512 to 64 degrades performance substantially, while $N_z{=}1$ with $D_z{=}512$ achieves near-peak results. This suggests that the cross-attention bottleneck must preserve sufficient representational capacity per latent, but the model can compress all spatial tokens into very few latent vectors without losing critical information. Adding refinement blocks ($B{=}2$) improves mAP for roof elements but slightly hurts mAP for roof materials; deeper refinement ($B{=}4$) degrades both tasks, likely due to over-smoothing. Attention rollout visualizations (\cref{fig:attention}) confirm that the model localizes target attributes across modalities, though for frequent classes like chimney, attention is diffuse (\cref{fig:attention-bad}), suggesting reliance on contextual shortcuts.

\begin{figure}[t]
  \centering
  \includegraphics[width=\linewidth]{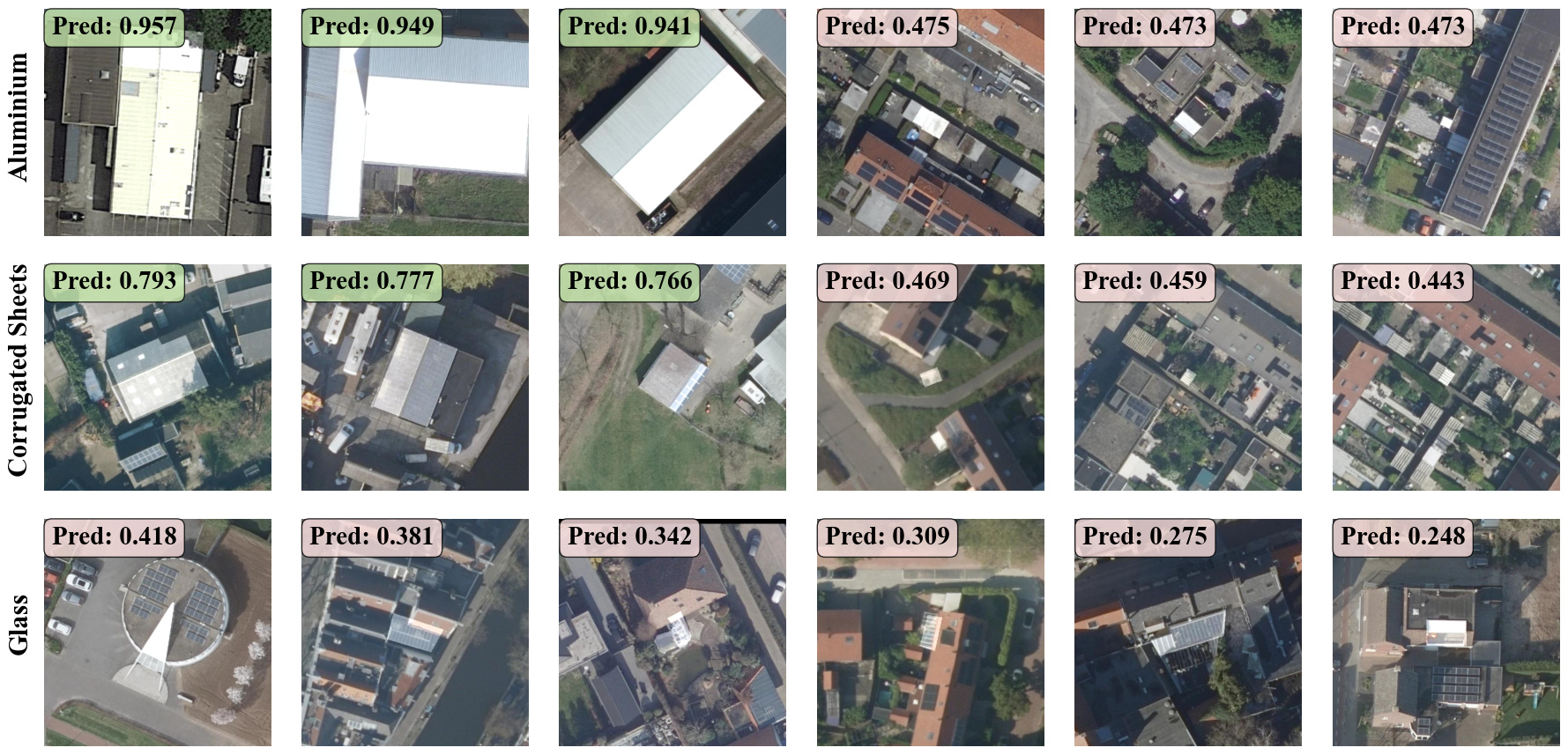}
  \caption{Challenging classes: Aluminium (top), Corrugated Sheets (middle), Glass (bottom). Left columns show top-3 correct predictions; right columns show top-3 incorrect predictions. These classes suffer from small sample sizes and confusion with visually similar materials.}
  \label{fig:uncommon}
\end{figure}

\paragraph{Limitations.}
The satellite-only model's higher macro mAP reflects a fundamental asymmetry: all 13 target classes are at least partially visible from above, but only a subset are informative from ground level. Street-level imagery also suffers from inconsistent coverage---some buildings have many high-quality views while others have none---and from noise sources including occlusion, privacy blurring, and camera positioning errors. Class imbalance (thatch: 184 samples, glass: 280) limits reliable evaluation for rare classes, and visually similar materials such as aluminium and corrugated sheets remain challenging regardless of the model (\cref{fig:uncommon}).

\section{Conclusion}
\label{sec:conclusion}

We presented a Perceiver IO-based architecture for multi-modal building inspection that fuses satellite and street-level imagery at the spatial token level using a shared DINOv2 backbone. Systematic experiments on a 32,135-building dataset across ten countries yield three main findings. First, appending the building footprint mask as a fourth input channel (RGB-M) provides the most effective strategy for isolating the target building, outperforming hard cropping and full-image baselines. Second, fully fine-tuned DINOv2-S consistently outperforms ResNet-50 on both modalities, even on this domain-specific dataset. Third, Perceiver IO fusion delivers substantial per-class improvements for attributes visible from street level (+11.3 AP for slate, +4.5 AP for glass, +1.3 AP for dormers), though the satellite-only baseline retains higher macro-averaged mAP due to classes that are only visible from above.

These results demonstrate that multi-modal fusion is most valuable when the modalities provide genuinely complementary views. Future work includes encoding street-level camera metadata (GPS coordinates, viewing angle, distance) as geometric positional embeddings to improve cross-view reasoning; exploiting the token-based architecture for patch-level masking, down-weighting or removing patches outside the building region for harder spatial attention than RGB-M; and multi-task learning that jointly performs classification, segmentation, and regression (e.g., building height, floor count) to encourage richer shared representations.

%%%%%%%%% SUPPLEMENTAL MATERIAL - unused figures for review

{
    \small
    \bibliographystyle{ieeenat_fullname}
    \bibliography{main}

@misc{eu_buildings,
  title={Energy performance of buildings directive},
  author={{European Commission}},
  year={2024},
  howpublished={\url{https://energy.ec.europa.eu/topics/energy-efficiency/energy-efficient-buildings/energy-performance-buildings-directive_en}}
}

@inproceedings{he2016deep,
  title={Deep residual learning for image recognition},
  author={He, Kaiming and Zhang, Xiangyu and Ren, Shaoqing and Sun, Jian},
  booktitle={Proceedings of the IEEE conference on computer vision and pattern recognition},
  pages={770--778},
  year={2016}
}

@article{oquab2023dinov2,
  title={Dinov2: Learning robust visual features without supervision},
  author={Oquab, Maxime and Darcet, Timoth{\'e}e and Moutakanni, Th{\'e}o and Vo, Huy and Szafraniec, Marc and Khalidov, Vasil and Fernandez, Pierre and Haziza, Daniel and Massa, Francisco and El-Nouby, Alaaeldin and others},
  journal={arXiv preprint arXiv:2304.07193},
  year={2023}
}

@article{ji2018fully,
  title={Fully convolutional networks for multisource building extraction from an open aerial and satellite imagery data set},
  author={Ji, Shunping and Wei, Shiqing and Lu, Meng},
  journal={IEEE Transactions on geoscience and remote sensing},
  volume={57},
  number={1},
  pages={574--586},
  year={2018},
  publisher={IEEE}
}

@inproceedings{zhao2018building,
  title={Building extraction from satellite images using mask R-CNN with building boundary regularization},
  author={Zhao, Kang and Kang, Jungwon and Jung, Jaewook and Sohn, Gunho},
  booktitle={Proceedings of the IEEE conference on computer vision and pattern recognition workshops},
  pages={247--251},
  year={2018}
}

@article{ji2019scale,
  title={A scale robust convolutional neural network for automatic building extraction from aerial and satellite imagery},
  author={Ji, Shunping and Wei, Shiqing and Lu, Meng},
  journal={International journal of remote sensing},
  volume={40},
  number={9},
  pages={3308--3322},
  year={2019},
  publisher={Taylor \& Francis}
}

@article{xu2019building,
  title={Building damage detection in satellite imagery using convolutional neural networks},
  author={Xu, Joseph Z and Lu, Wenhan and Li, Zebo and Khaitan, Pranav and Zaytseva, Valeriya},
  journal={arXiv preprint arXiv:1910.06444},
  year={2019}
}

@article{kim2021cnn,
  title={CNN algorithm for roof detection and material classification in satellite images},
  author={Kim, Jonguk and Bae, Hyansu and Kang, Hyunwoo and Lee, Suk Gyu},
  journal={Electronics},
  volume={10},
  number={13},
  pages={1592},
  year={2021},
  publisher={MDPI}
}

@article{wang2022building,
  title={Building extraction with vision transformer},
  author={Wang, Libo and Fang, Shenghui and Meng, Xiaoliang and Li, Rui},
  journal={IEEE Transactions on Geoscience and Remote Sensing},
  volume={60},
  pages={1--11},
  year={2022},
  publisher={IEEE}
}

@article{kaur2023large,
  title={Large-scale building damage assessment using a novel hierarchical transformer architecture on satellite images},
  author={Kaur, Navjot and Lee, Cheng-Chun and Mostafavi, Ali and Mahdavi-Amiri, Ali},
  journal={Computer-Aided Civil and Infrastructure Engineering},
  volume={38},
  number={15},
  pages={2072--2091},
  year={2023},
  publisher={Wiley Online Library}
}

@inproceedings{scheibenreif2022self,
  title={Self-supervised vision transformers for land-cover segmentation and classification},
  author={Scheibenreif, Linus and Hanna, Jo{\"e}lle and Mommert, Michael and Borth, Damian},
  booktitle={Proceedings of the IEEE/CVF Conference on Computer Vision and Pattern Recognition},
  pages={1422--1431},
  year={2022}
}

@article{lu2025vision,
  title={Vision foundation models in remote sensing: A survey},
  author={Lu, Siqi and Guo, Junlin and Zimmer-Dauphinee, James R and Nieusma, Jordan M and Wang, Xiao and Wernke, Steven A and Huo, Yuankai and others},
  journal={IEEE Geoscience and Remote Sensing Magazine},
  year={2025},
  publisher={IEEE}
}

@article{rundle2011using,
  title={Using Google Street View to audit neighborhood environments},
  author={Rundle, Andrew G and Bader, Michael DM and Richards, Catherine A and Neckerman, Kathryn M and Teitler, Julien O},
  journal={American journal of preventive medicine},
  volume={40},
  number={1},
  pages={94--100},
  year={2011},
  publisher={Elsevier}
}

@article{biljecki2021street,
  title={Street view imagery in urban analytics and GIS: A review},
  author={Biljecki, Filip and Ito, Koichi},
  journal={Landscape and Urban Planning},
  volume={215},
  pages={104217},
  year={2021},
  publisher={Elsevier}
}

@article{gonzalez2020automatic,
  title={Automatic detection of building typology using deep learning methods on street level images},
  author={Gonzalez, Daniela and Rueda-Plata, Diego and Acevedo, Ana B and Duque, Juan C and Ramos-Poll{\'a}n, Ra{\'u}l and Betancourt, Alejandro and Garc{\'\i}a, Sebastian},
  journal={Building and Environment},
  volume={177},
  pages={106805},
  year={2020},
  publisher={Elsevier}
}

@article{taoufiq2020hierarchynet,
  title={Hierarchynet: Hierarchical CNN-based urban building classification},
  author={Taoufiq, Salma and Nagy, Bal{\'a}zs and Benedek, Csaba},
  journal={Remote Sensing},
  volume={12},
  number={22},
  pages={3794},
  year={2020},
  publisher={MDPI}
}

@article{laupheimer2018neural,
  title={Neural networks for the classification of building use from street-view imagery},
  author={Laupheimer, Dominik and Tutzauer, Patrick and Haala, Norbert and Spicker, Marc},
  journal={ISPRS Annals of the Photogrammetry, Remote Sensing and Spatial Information Sciences},
  volume={4},
  pages={177--184},
  year={2018},
  publisher={Copernicus GmbH}
}

@inproceedings{zhao2018deep,
  title={Deep CNN-based methods to evaluate neighborhood-scale urban valuation through street scenes perception},
  author={Zhao, Junhan and Liu, Xiang and Kuang, Yanqun and Chen, Yingjie Victor and Yang, Baijian},
  booktitle={2018 IEEE third international conference on data science in cyberspace (dsc)},
  pages={20--27},
  year={2018},
  organization={IEEE}
}

@article{kong2020enhanced,
  title={Enhanced facade parsing for street-level images using convolutional neural networks},
  author={Kong, Gefei and Fan, Hongchao},
  journal={IEEE Transactions on Geoscience and Remote Sensing},
  volume={59},
  number={12},
  pages={10519--10531},
  year={2020},
  publisher={IEEE}
}

@article{iannelli2017extensive,
  title={Extensive exposure mapping in urban areas through deep analysis of street-level pictures for floor count determination},
  author={Iannelli, Gianni Cristian and Dell’Acqua, Fabio},
  journal={Urban Science},
  volume={1},
  number={2},
  pages={16},
  year={2017},
  publisher={MDPI}
}

@article{wang2024improving,
  title={Improving facade parsing with vision transformers and line integration},
  author={Wang, Bowen and Zhang, Jiaxin and Zhang, Ran and Li, Yunqin and Li, Liangzhi and Nakashima, Yuta},
  journal={Advanced Engineering Informatics},
  volume={60},
  pages={102463},
  year={2024},
  publisher={Elsevier}
}

@article{li2023semi,
  title={Semi-Supervised Learning from Street-View Images and OpenStreetMap for Automatic Building Height Estimation},
  author={Li, Hao and Yuan, Zhendong and Dax, Gabriel and Kong, Gefei and Fan, Hongchao and Zipf, Alexander and Werner, Martin},
  journal={arXiv preprint arXiv:2307.02574},
  year={2023}
}

@article{he2021urban,
  title={Urban neighbourhood environment assessment based on street view image processing: A review of research trends},
  author={He, Nan and Li, Guanghao},
  journal={Environmental Challenges},
  volume={4},
  pages={100090},
  year={2021},
  publisher={Elsevier}
}

@article{law2019take,
  title={Take a look around: using street view and satellite images to estimate house prices},
  author={Law, Stephen and Paige, Brooks and Russell, Chris},
  journal={ACM Transactions on Intelligent Systems and Technology (TIST)},
  volume={10},
  number={5},
  pages={1--19},
  year={2019},
  publisher={ACM New York, NY, USA}
}

@article{cao2018integrating,
  title={Integrating aerial and street view images for urban land use classification},
  author={Cao, Rui and Zhu, Jiasong and Tu, Wei and Li, Qingquan and Cao, Jinzhou and Liu, Bozhi and Zhang, Qian and Qiu, Guoping},
  journal={Remote Sensing},
  volume={10},
  number={10},
  pages={1553},
  year={2018},
  publisher={MDPI}
}

@article{hoffmann2019model,
  title={Model fusion for building type classification from aerial and street view images},
  author={Hoffmann, Eike Jens and Wang, Yuanyuan and Werner, Martin and Kang, Jian and Zhu, Xiao Xiang},
  journal={Remote Sensing},
  volume={11},
  number={11},
  pages={1259},
  year={2019},
  publisher={MDPI}
}

@article{srivastava2019understanding,
  title={Understanding urban landuse from the above and ground perspectives: A deep learning, multimodal solution},
  author={Srivastava, Shivangi and Vargas-Munoz, John E and Tuia, Devis},
  journal={Remote sensing of environment},
  volume={228},
  pages={129--143},
  year={2019},
  publisher={Elsevier}
}

@article{suel2021multimodal,
  title={Multimodal deep learning from satellite and street-level imagery for measuring income, overcrowding, and environmental deprivation in urban areas},
  author={Suel, Esra and Bhatt, Samir and Brauer, Michael and Flaxman, Seth and Ezzati, Majid},
  journal={Remote Sensing of Environment},
  volume={257},
  pages={112339},
  year={2021},
  publisher={Elsevier}
}

@article{huang2023comprehensive,
  title={Comprehensive urban space representation with varying numbers of street-level images},
  author={Huang, Yingjing and Zhang, Fan and Gao, Yong and Tu, Wei and Duarte, Fabio and Ratti, Carlo and Guo, Diansheng and Liu, Yu},
  journal={Computers, Environment and Urban Systems},
  volume={106},
  pages={102043},
  year={2023},
  publisher={Elsevier}
}

@article{fan2022multilevel,
  title={Multilevel spatial-channel feature fusion network for urban village classification by fusing satellite and streetview images},
  author={Fan, Runyu and Li, Jun and Li, Fengpeng and Han, Wei and Wang, Lizhe},
  journal={IEEE Transactions on Geoscience and Remote Sensing},
  volume={60},
  pages={1--13},
  year={2022},
  publisher={IEEE}
}

@article{chen2022multi,
  title={Multi-modal fusion of satellite and street-view images for urban village classification based on a dual-branch deep neural network},
  author={Chen, Boan and Feng, Quanlong and Niu, Bowen and Yan, Fengqin and Gao, Bingbo and Yang, Jianyu and Gong, Jianhua and Liu, Jiantao},
  journal={International Journal of Applied Earth Observation and Geoinformation},
  volume={109},
  pages={102794},
  year={2022},
  publisher={Elsevier}
}

@article{guo2024fusion,
  title={Fusion of satellite and street view data for urban traffic accident hotspot identification},
  author={Guo, Wentong and Xu, Cheng and Jin, Sheng},
  journal={International Journal of Applied Earth Observation and Geoinformation},
  volume={130},
  pages={103853},
  year={2024},
  publisher={Elsevier}
}

@article{xing2023flood,
  title={Flood vulnerability assessment of urban buildings based on integrating high-resolution remote sensing and street view images},
  author={Xing, Ziyao and Yang, Shuai and Zan, Xuli and Dong, Xinrui and Yao, Yu and Liu, Zhe and Zhang, Xiaodong},
  journal={Sustainable Cities and Society},
  volume={92},
  pages={104467},
  year={2023},
  publisher={Elsevier}
}

@misc{OpenStreetMap,
   author = {{OpenStreetMap contributors}},
   title = {{Planet dump retrieved from https://planet.osm.org }},
   howpublished = "\url{ https://www.openstreetmap.org }",
   year = {2025},
 }

@article{xie2021segformer,
  title={SegFormer: Simple and efficient design for semantic segmentation with transformers},
  author={Xie, Enze and Wang, Wenhai and Yu, Zhiding and Anandkumar, Anima and Alvarez, Jose M and Luo, Ping},
  journal={Advances in neural information processing systems},
  volume={34},
  pages={12077--12090},
  year={2021}
}

@inproceedings{cordts2016cityscapes,
  title={The cityscapes dataset for semantic urban scene understanding},
  author={Cordts, Marius and Omran, Mohamed and Ramos, Sebastian and Rehfeld, Timo and Enzweiler, Markus and Benenson, Rodrigo and Franke, Uwe and Roth, Stefan and Schiele, Bernt},
  booktitle={Proceedings of the IEEE conference on computer vision and pattern recognition},
  pages={3213--3223},
  year={2016}
}

@inproceedings{ilse2018attention,
  title={Attention-based deep multiple instance learning},
  author={Ilse, Maximilian and Tomczak, Jakub and Welling, Max},
  booktitle={International conference on machine learning},
  pages={2127--2136},
  year={2018},
  organization={PMLR}
}

@article{jaegle2021perceiver,
  title={Perceiver io: A general architecture for structured inputs \& outputs},
  author={Jaegle, Andrew and Borgeaud, Sebastian and Alayrac, Jean-Baptiste and Doersch, Carl and Ionescu, Catalin and Ding, David and Koppula, Skanda and Zoran, Daniel and Brock, Andrew and Shelhamer, Evan and others},
  journal={arXiv preprint arXiv:2107.14795},
  year={2021}
}
}

\end{document}